\documentclass[11pt]{article}

\usepackage[final]{acl}

\usepackage{times}
\usepackage{latexsym}

\usepackage[T1]{fontenc}

\usepackage[utf8]{inputenc}

\usepackage{microtype}

\usepackage{inconsolata}

\usepackage{graphicx}

\usepackage{hyperref}
\usepackage{url}
\usepackage{algorithm}
\usepackage{algorithmic}
\usepackage{multirow}
\usepackage{arydshln}
\usepackage{xcolor}
\usepackage{colortbl}
\usepackage{tikz}
\def\checkmark{\tikz\fill[scale=0.4](0,.35) -- (.25,0) -- (1,.7) -- (.25,.15) -- cycle;} 
\usepackage{amsthm}
\usepackage{amsmath}
 \usepackage{amsfonts}
\theoremstyle{definition}
\newtheorem{definition}{Definition}
\newtheorem{theorem}{Theorem}
\newtheorem{lemma}[theorem]{Lemma}
\usepackage{booktabs}
\usepackage{caption}

\title{Privacy-Preserving Generation of Clinical Narratives from Medical Terminologies}

\author{
Yuping Wu\textsuperscript{1}, 
Viktor Schlegel\textsuperscript{1,2,3}\footnotemark[2], 
Warren Del-Pinto\textsuperscript{1}, 
Srinivasan Nandakumar\textsuperscript{2},
Iqra Zahid\textsuperscript{2},\\
\textbf{Yidan Sun\textsuperscript{2},}
\textbf{Hao Li\textsuperscript{3},}
\textbf{Usama Farghaly Omar\textsuperscript{4},} 
\textbf{Amirah Jasmine\textsuperscript{3},}
\textbf{Arun-Kumar Kaliya-Perumal\textsuperscript{5},}\\
\textbf{Chun Shen Tham\textsuperscript{1},}
\textbf{Gabriel Connors\textsuperscript{1},}
\textbf{Anil Anthony Bharath\textsuperscript{2,3},}
\textbf{Goran Nenadic\textsuperscript{1}}\\
$^{1}$Univeristy of Manchester, Department of Computer Science \\
$^{2}$Imperial Global Singapore \quad 
$^{3}$Imperial College London, Department of Bioengineering \\
$^{4}$Khoo Teck Puat Hospital, Orthopaedic Surgery Department, Singapore \\
$^{5}$Nanyang Technological University, Rehabilitation Research Institute of Singapore\\
}

\begin{document}
\maketitle
\renewcommand{\thefootnote}{\fnsymbol{footnote}} 
\footnotetext[2]{Correspondence: \href{mailto:v.schlegel@imperial.ac.uk}{v.schlegel@imperial.ac.uk}}
\begin{abstract}
In high-stakes domains such as healthcare, privacy concerns severely limit the use of real-world training data. Differentially private (DP) synthetic data offers a promising alternative with formal privacy guarantees, but achieving strong utility remains challenging for clinical note generation due to domain specificity and long-form text complexity.
We present \textbf{Term2Note}\footnote{Code: \url{https://github.com/yuping-wu/Term2Note}}, a method for synthesising full-length clinical notes under DP constraints. By structurally separating content and form, Term2Note generates section-wise note content conditioned on medical terms, with terms and notes privatised under separate DP constraints, and applies a DP quality maximiser to improve outputs.
Experiments demonstrate that Term2Note produces synthetic notes with statistical properties closely aligned with real clinical notes, and that downstream models trained on these notes achieve performance comparable to those trained on real clinical data. Compared to existing DP text generation baselines, Term2Note substantially improves both fidelity and utility, without relying on label distribution assumptions, highlighting its effectiveness as a practical privacy-preserving alternative to real clinical notes.
\end{abstract}

\section{Introduction}
The scaling law of neural language models \citep{DBLP:journals/corr/abs-2001-08361} suggests that model performance improves with larger training corpora. As large language models (LLMs) continue to scale in size, with models such as Llama 3 \citep{llama3model}, Gemma 3 \citep{gemma3model}, and Qwen3 \citep{DBLP:journals/corr/abs-2505-09388} ranging from 0.6B to over 405B parameters, the demand for large-scale, high-quality training data has risen accordingly. Synthetic data generation with LLMs has therefore emerged as a promising approach, with instruction-following synthetic datasets \citep{DBLP:conf/emnlp/SchickS21a, alpaca, DBLP:journals/corr/abs-2503-02445} proving effective for model pretraining and fine-tuning. This is especially relevant for high-stakes domains such as healthcare \citep{DBLP:journals/corr/abs-2506-07584}, where real data is sensitive, highly regulated, and difficult to share~\citep{DBLP:journals/corr/abs-2503-20846,sun2025evaluating}. Although large amounts of clinical data exist within healthcare institutions, access to these datasets remains extremely limited. A practical and privacy-conscious solution is to share synthetic versions of sensitive clinical data instead of the raw data itself. However, because even synthetic texts can leak private information~\cite{pmlr-v267-meeus25a,sun2025synbench}, to make such sharing viable, formal privacy guarantees are essential.

Differential privacy (DP) provides a principled framework for this purpose \citep{DBLP:journals/ejisec/AlzoubiM25}. By bounding the influence of any individual record on the synthesised dataset, DP offers quantifiable privacy guarantees. Prior work on DP-based text generation has focused mainly on short-form texts in low-risk domains such as reviews \citep{DBLP:conf/acl/YueILKMS0LS23, DBLP:journals/corr/abs-2306-01684, DBLP:conf/emnlp/MatternJWSS22, DBLP:conf/acl/FlemingsA24}. In biomedical settings, efforts have been restricted to synthesising (public) PubMed abstracts \citep{DBLP:conf/icml/Xie0BGYINJZL0Y24} and relatively short clinical passages \citep{DBLP:journals/health/AzizAFJYM22, DBLP:conf/emnlp/RameshGMBPF24}, with no prior work addressing the more complex task of generating full-length clinical notes under DP constraints.

Synthesising clinical notes with DP presents two key challenges. First, \textit{generation complexity} \citep{DBLP:conf/nips/KweonKKCYKYWC24, weetman2021makes}: clinical notes are long and exhibit diverse structures and free-form content, making it more difficult for generative models to maintain coherence and quality, particularly under privacy constraints. Second, \textit{domain specificity} \citep{adnan2010assessing}: clinical notes typically contain extensive domain-specific terminologies, which require expert knowledge to understand and reproduce accurately.

\begin{figure*}[t]
    \centering
        \includegraphics[width=0.95\textwidth]{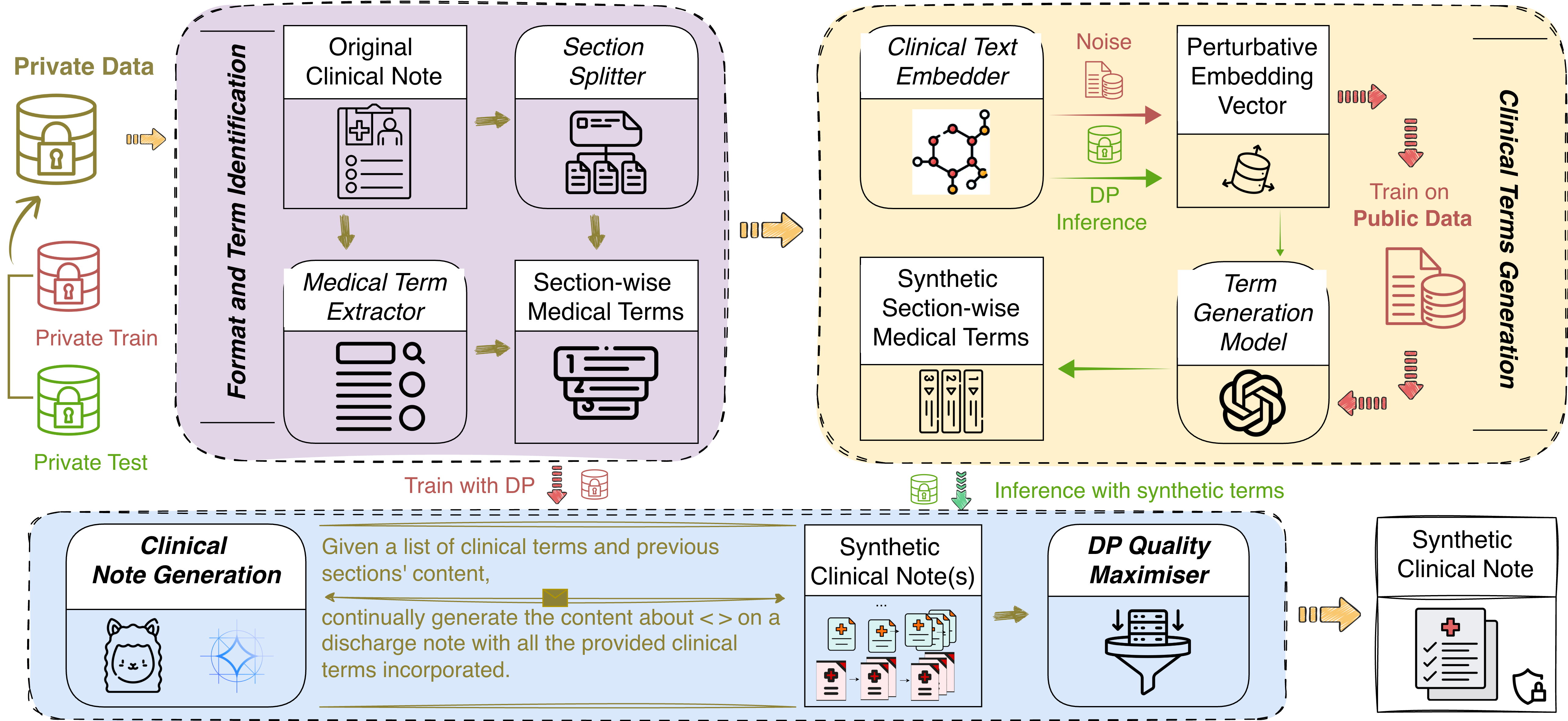}
    \caption{Overview of \textbf{Term2Note}. 
            An original clinical note is split into sections and associated medical terms, which are embedded and optionally privatised via DP perturbation. A term generation model produces synthetic terms from these embeddings, and a DP-trained note generation model synthesises section-wise notes conditioned on them. A DP quality maximiser then selects the final synthetic note. Appendix \ref{appendix:example_demo} illustrates the whole process with a concrete example.
            }
    \label{fig:model}
\end{figure*}

In this paper, we tackle this underexplored and challenging task by proposing Term2Note, a novel methodology for DP synthetic clinical note generation. As illustrated in Figure \ref{fig:model}, Term2Note addresses the challenge of long-form generation by leveraging domain-specific document structures to decompose the task into smaller, section-wise subtasks. To handle domain specificity, generation is conditioned not on generic metadata (e.g., class labels or diagnosis codes such as \textit{Enterocolitis due to Clostridium difficile}), but on salient clinical terms (e.g., \textit{[diarrhea, Clostridium difficile colitis, Vancomycin]}), which are subjected to an additional layer of DP protection as rare combinations of such terms might bear patient-identifying information. This term-based conditioning strategy allows the model to generate text that is both clinically meaningful and structurally coherent, while also enabling more fine-grained privacy control.

Our experimental results show that Term2Note: (1) produces synthetic notes with high structural and semantic fidelity to real clinical data; (2) provides high utility for downstream tasks, such as ICD code prediction; (3) satisfies formal DP guarantees, facilitating data sharing with quantifiable privacy protection. Moreover, Term2Note consistently demonstrates stronger performance than baselines across a wide range of evaluation metrics.
In summary, Term2Note offers a promising solution for privacy-preserved clinical notes sharing.

\section{Background \& Related Work}

De-identification is widely used to protect sensitive data by removing explicit personal identifiers. However, it does not provide formal privacy guarantees, and individuals can often be re-identified through quasi-identifiers or linkage with external datasets. For example, users in the anonymised Netflix dataset were successfully re-identified by linking it with IMDb ratings \cite{narayanan2006break}. In contrast, differential privacy (DP) provides formal, quantifiable privacy guarantees against such inference attacks. We briefly review the theoretical foundations of DP below.

\begin{definition} \citep{DBLP:conf/tcc/DworkMNS06}
     A randomised algorithm $\mathcal{M}: \mathbb{N}^{|\mathcal{X}|} \to R$ is said to be \emph{$(\epsilon, \delta)$-\textbf{differentially private (DP)}}, if, for any two neighboring datasets $D$ and $D'$ differing in one single instance, and for all subsets $S$ of the output space of $\mathcal{M}$, it has \mbox{$\mathbb{P}[\mathcal{M}(D) \in S] \leq e^\epsilon \mathbb{P}[\mathcal{M}(D')\in S] + \delta$}.
\end{definition}
The definition implies the probability distributions induced by $\mathcal{M}$ on neighboring datasets must be close, with their likelihood ratios bounded by a multiplicative factor of $e^\epsilon$ and an additive slack of $\delta$.  
Smaller $\epsilon$ values indicate stronger privacy, while $\delta$ denotes the (typically negligible) probability of a privacy breach exceeding the $e^\epsilon$ bound. 

\begin{theorem}(\textbf{Post-Processing}) \citep{DBLP:journals/fttcs/DworkR14}
    Let $\mathcal{M}: \mathbb{N}^{|\mathcal{X}|} \to R$ be a randomised algorithm that is $(\epsilon, \delta)$-DP. Let $f: R \to R'$ be an arbitrary randomised function. Then \mbox{$f \circ \mathcal{M}: \mathbb{N}^{|\mathcal{X}|} \to R'$} is also $(\epsilon, \delta)$-DP.
\end{theorem}

\begin{theorem}(\textbf{Parallel Composition}) \citep{DBLP:conf/sigmod/McSherry09}
    Let dataset $D=D_1 \cup \cdots \cup D_k$, and $D_i \cap D_j = \emptyset$ for $i \neq j$. Let $\mathcal{M}_i: \mathbb{N}^{|\mathcal{X}_i|} \to R_i$ be a randomised algorithm that is $(\epsilon_i, \delta_i)$-DP, for $i \in [k]$. Then $\mathcal{M}(D)=(\mathcal{M}_1(D_1), \cdots, \mathcal{M}_k(D_k))$ is $(\max_i \epsilon_i, \max_i\delta_i)$-DP.
\end{theorem}

The post-processing property states that applying any function to the output of an $(\epsilon, \delta)$-DP algorithm preserves the same privacy guarantee. Thus, a DP-trained generative model remains private during downstream use.
The parallel composition property states that applying DP mechanisms to disjoint data subsets results in an overall privacy loss bounded by the maximum individual $(\epsilon, \delta)$ values, enabling clinical terms and notes to be privatised independently in our method.

\textbf{DP in Deep Learning} is commonly achieved by adding random noise, typically Gaussian noise, during training. DP-SGD \citep{DBLP:conf/ccs/AbadiCGMMT016} enforces DP by clipping per-sample gradients and injecting noise at each optimisation step, limiting the influence of individual training examples on model parameters. Subsequent work has improved the efficiency of DP-SGD \citep{DBLP:journals/corr/abs-2009-03106, DBLP:conf/icml/BuWZK23, DBLP:journals/corr/abs-2109-12298} by reducing the time and memory overhead associated with per-example gradient computation. Unless otherwise specified, we adopt FastDP \citep{DBLP:conf/icml/BuWZK23} for DP fine-tuning throughout this paper.

\textbf{Synthetic text generation} has progressed rapidly with the rise of LLMs, especially through instruction-following datasets that enhance downstream performance \citep{alpaca, DBLP:journals/corr/abs-2304-03277, DBLP:conf/emnlp/SchickS21a}. Generating synthetic clinical text, however, is substantially more challenging due to the domain complexity, expert knowledge requirements, and privacy concerns. Prior attempts have often produced datasets with limited downstream utlity \citep{DBLP:conf/bionlp/LiWSBNKZB0N23, DBLP:conf/clef/SchlegelLW0NKBZ23}. Recent work has sought to improve realism by prompting LLMs with control code (e.g., ICD codes) \citep{DBLP:journals/jamia/FalisGDDBHPBA24} or transforming biomedical abstracts (e.g., PubMed content) into clinical-style text \citep{DBLP:conf/acl/KweonKKICBOLMYB24}. While promising, these methods do not address privacy, a central concern in clinical settings. A straightforward strategy is de-identifying private information and prompting LLMs to fill in the gaps \citep{sarkar2025not}, but a more principled approach is to fine-tune or instruction-tune generative models with DP for formal protection \citep{DBLP:journals/health/AzizAFJYM22, DBLP:conf/emnlp/RameshGMBPF24, DBLP:journals/corr/abs-2409-07809}. However, prior studies operate on relatively short clinical texts, either naturally brief or deliberately truncated, making generation considerably easier. In contrast, DP-constrained generation of full-length clinical notes, as addressed in this paper, is more challenging due to their length, unstructured format, and high variability.

\section{Methodology}

\subsection{Problem Statement}
Given a private dataset $D^{\text{src}}$ consisting of clinical notes, our goal is to develop a mechanism $\mathcal{M}$ that satisfies $(\epsilon, \delta)$-DP and produces a synthetic dataset $D^{\text{syn}}$
as its output. 
Let $X^{\text{src}} \in D^{\text{src}}$ denote an original clinical note and $X^{\text{syn}} \in D^{\text{syn}}$ its corresponding synthetic version. 
To support the development and evaluation of the mechanism, we partition the private dataset into a training set $D^{\text{src}}_\text{train}$ and a test set $D^{\text{src}}_\text{test}$. The training set is used to develop $\mathcal{M}$, while the test set remains completely unseen by $\mathcal{M}$ to provide an unbiased evaluation. Additionally, we assume access to a public dataset of clinical terms, denoted as $D_\text{public}$, which can be automatically derived from publicly available medical resources and therefore used freely without privacy constraints.

We introduce \textbf{Term2Note}, a section-wise DP generation framework for clinical notes. An overview is illustrated in Figure \ref{fig:model}, with the detailed procedure in Algorithm \ref{algo:Term2Note}. In the following, we first elaborate on Term2Note alongside the algorithm, and then present the implementation details.

\subsection{Format and Term Identification}
\label{sec:format_and_term}
Since our framework synthesises clinical notes via section-wise generation conditioned on medical terms, we first standardise the structure of the notes and identify salient clinical terms. Let \textsc{SecSplit} denote an automatic section segmentation module. Given an original clinical note $X^{\text{src}}$, \textsc{SecSplit} outputs a list of $m$ segmented sections: $[\text{SEC}_1^{\text{src}}, ..., \text{SEC}_m^{\text{src}}] = \textsc{SecSplit}(X^{\text{src}})$, where $\text{SEC}_i^{\text{src}} = $ ``'' if the $i$-th section is absent. Next, we apply an automatic term extraction module, denoted as \textsc{TermExt}, to each section to identify clinically salient terms, resulting in a list of section-specific medical terms: $[T_1^{\text{src}}, ..., T_m^{\text{src}}] = [\textsc{TermExt}(\text{SEC}_1^{\text{src}}), ..., \textsc{TermExt}(\text{SEC}_m^{\text{src}})]$.

\subsection{Clinical Terms Generation}
\label{sec:term_gen}
Clinical terms from private notes may still contain sensitive information. For example, unique combinations of diagnoses and procedures could re-identify patients. A uniqueness analysis on a large subset of the MIMIC-IV dataset reveals that (1) although clinical terms are often shared across notes, more than half of the notes contain at least one clinical term that is unique to that note, and (2) certain sections, such as clinical course and medication, are highly distinctive, with over 96\% of notes containing unique combination of terms in these sections. The presence of note-specific clinical terms increases the risk of patient re-identification and privacy exposure. To mitigate this risk, we introduce an optional DP step for term generation. We formulate it as a reconstruction task, fine-tuning a generative model on the public term dataset $D_\text{public}$ to recover term lists from embeddings, and then applying it to private data under DP constraints.

Specifically, any given (section-wise) term list is first embedded using a clinical text embedder, denoted \textsc{Emb}. A projection layer \textsc{Proj}, parameterised by $\theta_p$, maps the embeddings to the hidden dimensionality required by the generative model \textsc{TermGen}, parameterised by $\theta_t$. The model then reconstructs the original term list from the projected embeddings. To ensure privacy, we adapt the DPRP schema \citep{DBLP:conf/uai/Gondara020}, a model-agnostic DP mechanism originally proposed for tabular data, to term embeddings derived from private notes, denoted as $\textsc{DPRP}^*$. The procedure perturbs private embeddings in four steps: (1) add dimension-wise random noise to input embeddings; (2) compute the covariance matrix of the input embeddings and add random noise; (3) perform singular value decomposition (SVD) on the noisy covariance matrix; (4) reconstruct the inputs from the noisy embeddings and the right singular vectors. The pseudocode of $\textsc{DPRP}^*$ is provided in Appendix \ref{appendix:dprp_algo}. To minimise the distribution difference between embeddings used in training and those perturbed during inference, we additionally add Gaussian noise to the embeddings during training. 
Formally, the process is defined as follows:
\begin{gather}
    E' = 
    \begin{cases}
        \textsc{Emb}(D_{\text{public}}) + \mathcal{N}(0, \sigma_{\text{emb}}), & \text{if training},\\
        \textsc{DPRP}^*(\textsc{Emb}(T^{\text{src}}), \frac{\epsilon_t}{m}, \frac{\sigma_t}{m}), & \text{otherwise}.
    \end{cases} \\
    T^{\text{syn}} \sim \textsc{TermGen}_{\theta_t}(\textsc{Proj}_{\theta_{p}}(E'))
\end{gather}
Here, $\epsilon_t$ and $\sigma_t$ denote the privacy parameters of $\textsc{DPRP}^*$, and $m$ is the number of sections in a single note. Since $E'$ is computed at the section level, the overall privacy cost for an entire note accumulates across sections. To account for this, we distribute the privacy budget evenly by scaling the cost for each section to $\tfrac{1}{m}$ of the total budget.

\subsection{Clinical Note Generation}
\label{sec:note_gen}
We define section-wise clinical note generation as a conditional text generation task. Given the task instruction $I$, the content of the previous (generated) sections $[\text{SEC}_1, ..., \text{SEC}_{i-1}]$, and a list of clinical terms $T_i$ for the current section, a generative model \textsc{NoteGen}, parameterised by $\theta_n$, is trained to produce the $i$-th section of the note under $(\epsilon_n, \delta_n)$-DP. During inference, each section is sampled sequentially as:
\begin{equation}
    \text{SEC}_i^{\text{syn}} \sim \textsc{NoteGen}_{\theta_n}(I, [\text{SEC}_1^{\text{syn}}, ..., \text{SEC}_{i-1}^{\text{syn}}], T_i)
\end{equation}
Here, $T_i$ can be the original extracted terms $T_i^{\text{src}}$ or the synthetic privatised terms $T_i^{\text{syn}}$, depending on the privacy configuration. With the section group name provided, the instruction $I$ is defined as shown in Figure \ref{fig:model}.
Finally, a synthetic full note is obtained by concatenating the generated sections: $X^{\text{syn}} = [\text{SEC}_1^{\text{syn}}, ..., \text{SEC}_m^{\text{syn}}]$.

\subsection{DP Quality Maximiser}
\label{sec:dp_ranker}
To improve the quality of the synthetic data, we introduce a quality maximisation strategy during inference by leveraging the generative capabilities of LLMs. Specifically, instead of generating a single synthetic note, we perform preference sampling on $k$ candidate notes for $X^{\text{src}}$, denoted $X^{\text{syn}}[1:k]$. Notably, this sampling procedure preserves the DP guarantee due to the post-processing property of DP.
To select the most fluent and coherent output among the candidates, we use perplexity as the preference model. Perplexity reflects the likelihood of a sequence under an LLM, computed as the exponentiated average negative log-likelihood of the tokens. Lower perplexity indicates higher linguistic plausibility. To avoid bias from the generator itself, we compute perplexity scores using a reference domain-specialised LLM, denoted $\textsc{LLM}_{\text{ppl}}$. This ensures a more objective assessment of sequence quality.
Formally, the perplexity of a candidate note $X^{\text{syn}}[i]$ is given by $\text{PPL}(X^{\text{syn}}[i]) = \exp \left(-\frac{1}{t}\sum_{i=1}^{t}log \;\textsc{LLM}_{\text{ppl}}(d_i|d_{<i})\right)$,
where $d_{1, \dots, t}$ are tokens in $X^{\text{syn}}[i]$.
The synthetic note with the lowest perplexity score is selected.

\subsection{Privacy Analysis} 
The overall privacy guarantee of Term2Note depends on the composition of its two DP components: \textsc{TermGen} and \textsc{NoteGen}. Specifically, \textsc{TermGen} is trained on $D_{\text{public}}$ and can optionally be applied to privatise terms for $D_{\text{test}}^{\text{src}}$, while \textsc{NoteGen} is trained on $D_{\text{train}}^{\text{src}}$ under DP. Since \textsc{TermGen} and \textsc{NoteGen} operate on disjoint subsets of private data, the overall privacy loss can be computed using parallel composition.  Formally, the total privacy guarantee $(\epsilon, \delta)$ is defined as below, and the proof is provided in Appendix \ref{appendix:privacy_proof}.
\begin{equation}
    (\epsilon, \delta) = (\max(\epsilon_n, \epsilon_t), \max(\delta_n, \delta_t))
\end{equation}

\subsection{Implementation Details}
\paragraph{\textsc{SecSplit}} 
Clinical notes lack a universally standardised structure, and while formats such as SOAP are common, they often require manual annotation for accurate segmentation \citep{DBLP:conf/coling/GaoDMXCA22}. To enable automatic sectioning, we analyse formatting patterns in the raw notes and design a rule-based segmentation approach using regular expressions to identify section titles. Section spans are determined greedily, from each detected title to the next. 
We automatically curate a list of frequent section titles and group them into six semantic categories: \textit{Patient Information}, \textit{Clinical Course \& History}, \textit{Examinations \& Findings}, \textit{Laboratory \& Imaging Results}, \textit{Hospital Stay \& Treatment}, and \textit{Medications \& Discharge Plan}. This taxonomy underpins \textsc{SecSplit}, which splits each note into up to six standardised sections. The full list of section titles and groupings is provided in Appendix~\ref{appendix:section_grouping}.

\paragraph{\textsc{TermExt}} 
Biomedical vocabularies vary by taxonomy. In this work, we focus on terms from SNOMED CT, a comprehensive clinical vocabulary widely used in electronic health records (EHRs) and included in the Unified Medical Language System (UMLS) metathesaurus \citep{UMLS2025}.
To extract medical terms from clinical text, we use QuickUMLS \citep{Soldaini2016QuickUMLSAF}, an unsupervised approximate string-matching tool over UMLS concepts, and retain only SNOMED CT terms.

\paragraph{Backbone Models} 
For clinical term embeddings, we use the encoder-only model MedEmbed-large \citep{balachandran2024medembed} as the embedder \textbf{\textsc{Emb}}, which is fine-tuned for medical and clinical text. 
For the two generative modules in Term2Note, we adopt lightweight yet effective language models: GPT2-Large \citep{radford2019language}  for the term generator \textbf{$\textsc{TermGen}$}, and Llama-3.2-1B \citep{llama3model} or Gemma-3-1B \citep{gemma3model} for the note generator \textbf{$\textsc{NoteGen}$}. Using both Llama and Gemma increases architectural diversity and reduces model-specific bias while enabling reproducible evaluation.
To compute the perplexity of generated notes, we use Asclepius-Llama3-8B \citep{DBLP:conf/acl/KweonKKICBOLMYB24} as the reference model \textbf{$\textsc{LLM}_{\text{ppl}}$}. Pre-trained on clinical text and supporting up to 8192 tokens as input, it mitigates domain mismatch and accommodates full-length clinical notes.

\begin{table*}[h]
    \centering
    \resizebox{\textwidth}{!}{

    \setlength{\tabcolsep}{1mm} 
    \small
    \begin{tabular}{ll c l cc l c c| cc l cc l cc}
        \toprule
        & \multirow{3}{*}{\textbf{Method}} 
        & \multicolumn{6}{c}{\textbf{Fidelity}}
        & & \multicolumn{8}{c}{\textbf{Utility}} \\
        \cmidrule{3-8}\cmidrule{10-17}
        & 
        & \textbf{Length} 
        & 
        & \multicolumn{2}{c}{\textbf{Unary/Binary Term}} & 
        & \textbf{Semantic} &
        & \multicolumn{2}{c}{\textbf{F1}}  &
        & \multicolumn{2}{c}{\textbf{AUC}}  &
        & \multicolumn{2}{c}{\textbf{Precision@$k$}} \\
        \cmidrule{3-3}\cmidrule{5-6}\cmidrule{8-8}
        \cmidrule{10-11}\cmidrule{13-14}\cmidrule{16-17}
        & & \textbf{KL Div.}$\downarrow$ &
        & \textbf{Jaccard}$\uparrow$ & \textbf{KL Div.}$\downarrow$ & 
        & \textbf{MAUVE}$\uparrow$  &
        & \textbf{Micro} & \textbf{Macro} &
        &  \textbf{Micro} & \textbf{Macro} &
        &  $\mathbf{k=3}$ & $\mathbf{k=5}$  \\
        \midrule
        & {Original Data} 
        & & & & & & & 
        & 57.03 & 30.80 & & 82.01 & 58.88 & & 68.93 & 62.14 \\
        \midrule
        \multirow{32}{*}{\rotatebox{90}{Synthetic}} 
        & \multicolumn{16}{c}{$\epsilon=\infty$} \\ 
        \cmidrule(lr{0.001pt}){2-17}
        & AUG-PE $(\epsilon_\text{meta}=\infty, \epsilon_n=\infty)$
        & 11.96 & & 0.14/0.02 & 7.59/16.34 & & 0.01 & 
        & 45.82 & 14.84 & & 79.52 & 54.35 & & 68.48 & 61.77 \\
        & FastDP $(\epsilon_\text{meta}=\infty, \epsilon_n=\infty)$
        & 1.04 & & \textbf{\underline{0.53/0.28}} & 0.32/\textbf{\underline{0.84}} & & 0.12 & 
        & \underline{53.02} & \underline{25.51} & & 79.35 & 51.05 & & \underline{69.77} & 60.39 \\
        \addlinespace
        & Term2Note $(\epsilon_\text{meta}=\infty, \epsilon_n=\infty)$
        & \underline{0.25} & & 0.52/0.20 & \textbf{\underline{0.22}}/1.08 & & \textbf{\underline{0.59}} & 
        & 49.95 & 21.89 & & \textbf{\underline{81.40}} & \textbf{\underline{55.43}} & & \underline{69.77} & \underline{61.96} \\
        & Term2Note $(\epsilon_t=\infty, \epsilon_n=\infty)$
        & 0.68 & & 0.43/0.14 & 0.45/1.95 & & 0.46 & 
        & 49.24 & 21.90 & & 80.24 & 51.54 & & 67.81 & 61.08 \\
        & Term2Note $(\epsilon_t=\infty, \epsilon_n=8)$
        & 0.26 & & 0.39/0.13 & 0.50/1.19 & & 0.35 & 
        & 48.16 & 20.63 & & 78.72 & 50.01 & & 65.35 & 58.72 \\
        & \multicolumn{1}{l}{Term2Note ($\epsilon_t=\infty, \epsilon_n=5$)}
        & 0.28 & & 0.38/0.13 & 0.53/1.23 & & 0.27 & 
        & 51.00 & 22.73 & & 78.80 & 50.37 & & 67.15 & 59.91 \\
        & \multicolumn{1}{l}{Term2Note ($\epsilon_t=\infty, \epsilon_n=2$)}
        & 0.43 & & 0.37/0.12 & 0.60/1.35 & & 0.36 & 
        & 48.57 & 20.31 & & 79.56 & 51.75 & & 68.64 & \underline{60.41} \\
        \addlinespace
        \cmidrule(lr{0.001pt}){2-17}
        & \multicolumn{16}{c}{$\epsilon=8$} \\
        \cmidrule(lr{0.001pt}){2-17}
        & AUG-PE $(\epsilon_\text{meta}=\infty, \epsilon_n=8)$
        & 11.71 & & 0.19/0.03 & 5.03/12.18 & & 0.01 & 
        & 40.73 & 13.28 & & 78.13 & \underline{53.49} & & 63.24 & 59.03 \\
        & FastDP $(\epsilon_\text{meta}=\infty, \epsilon_n=8)$
        & 4.51 & & 0.31/0.10 & 2.88/5.88 & & 0.02 & 
        & 48.58 & 16.40 & & \underline{80.74} & 51.57 & & \textbf{\underline{69.79}} & \underline{61.59} \\
        \addlinespace
        & Term2Note $(\epsilon_\text{meta}=\infty, \epsilon_n=8)$
        & 0.39 & & \underline{0.40/0.13} & \underline{0.47/1.14} & & \underline{0.53} & 
        & 49.71  & 21.28 & & 80.03 & 52.80 & & 67.49 & 61.48 \\
        & \multicolumn{1}{l}{Term2Note ($\epsilon_t=8, \epsilon_n=8$)}
        & 0.16 & & 0.38/0.13 & 0.62/1.15 & & 0.37 & 
        & 52.31 & \underline{26.50} & & 78.19 & 50.17 & & 67.81 & 57.36 \\
        & \multicolumn{1}{l}{Term2Note ($\epsilon_t=5, \epsilon_n=8$)}
        & \underline{\textbf{0.15}} & & 0.39/0.13 & 0.64/1.16 & & 0.46 & 
        & 49.52 & 21.36 & & 79.08 & 50.11 & & 68.29 & 60.68 \\
        & \multicolumn{1}{l}{Term2Note ($\epsilon_t=2, \epsilon_n=8$)}
        & 0.19 & & 0.38/0.12 & 0.61/1.18 & & 0.38 & 
        & \underline{53.25} & 24.66 & & 78.85 & 49.41 & & 68.31 & 59.91 \\
        \addlinespace
        \cmidrule(lr{0.001pt}){2-17}
        & \multicolumn{16}{c}{$\epsilon=5$} \\
        \cmidrule(lr{0.001pt}){2-17}
        & AUG-PE $(\epsilon_\text{meta}=\infty, \epsilon_n=5)$
        & 11.70 & & 0.11/0.01 & 7.75/13.58 & & 0.01 & 
        & 48.10 & 17.44 & & 77.78 & 53.33 & & 63.01 & 56.94 \\
        & FastDP $(\epsilon_\text{meta}=\infty, \epsilon_n=5)$
        & 3.40 & & 0.29/0.09 & 2.97/5.67 & & 0.04 & 
        & 49.30 & 16.23 & & \underline{80.54} & \underline{54.22} & & \underline{67.31} & \textbf{\underline{61.98}} \\
        \addlinespace
        & Term2Note $(\epsilon_\text{meta}=\infty, \epsilon_n=5)$
        & \underline{0.20} & & \underline{0.41/0.14} & \underline{0.42/1.17} & & 0.39 & 
        & 47.94 & 20.31 & & 79.29 & 51.19 & & 66.04 & 61.69 \\
        & \multicolumn{1}{l}{Term2Note ($\epsilon_t=5, \epsilon_n=5$)}
        & \underline{0.20} & & 0.38/0.13 & 0.63/1.19 & & \underline{0.43} & 
        & \textbf{\underline{54.83}} & \textbf{\underline{28.96}} & & 78.20 & 50.32 & & 64.56 & 57.18 \\
        & \multicolumn{1}{l}{Term2Note ($\epsilon_t=2, \epsilon_n=5$)}
        & 0.36 & & 0.38/0.12 & 0.64/1.32 & & 0.36 & 
        & 51.26 & 21.45 & & 79.05 & 50.44 & & 66.36 & 60.49 \\
        \addlinespace
        \cmidrule(lr{0.001pt}){2-17}
        & \multicolumn{16}{c}{$\epsilon=2$} \\
        \cmidrule(lr{0.001pt}){2-17}
        & AUG-PE $(\epsilon_\text{meta}=\infty, \epsilon_n=2)$
        & 12.11 & & 0.19/0.03 & 5.42/11.85 & & 0.01 & 
        & 40.90 & 13.57 & & 78.29 & \underline{53.38} & & 63.74 & 60.10 \\
        & FastDP $(\epsilon_\text{meta}=\infty, \epsilon_n=2)$
        & 9.67 & & 0.14/0.03 & 5.79/10.89 & & 0.01 & 
        & 51.06 & 20.04 & & \underline{79.98} & 51.31 & & 66.32 & 59.99 \\
        \addlinespace
        & Term2Note $(\epsilon_\text{meta}=\infty, \epsilon_n=2)$
        & \underline{0.43} & & \underline{0.39/0.12} & \underline{0.48/1.17} & & \underline{0.31} & 
        & 51.78 & \underline{23.36} & & 79.00 & 50.60 & & 67.00 & 59.52 \\
        & \multicolumn{1}{l}{Term2Note ($\epsilon_t=2, \epsilon_n=2$)}
        & 0.48 & & 0.37/0.12 & 0.68/1.29 & & \underline{0.31} & 
        & \underline{51.87} & 23.06 & & 79.43 & 51.30 & & \underline{69.45} & 60.31 \\
        \addlinespace
        \bottomrule
    \end{tabular}
    }
    \setlength{\tabcolsep}{6pt}  

    \caption{Fidelity and utility evaluation of synthetic datasets generated by different methods using Llama-3.2-1B. Term2Note with $\epsilon_{\text{meta}}=\infty$ indicates that note generation is conditioned on ground-truth labels, i.e., without perturbing the input labels, whereas $\epsilon_t$ denotes privatised input labels. Appendix~\ref{appendix:privacy_accounting} summarises the privacy budget allocation for the experiments in this table. Utility metrics are averaged across 5-fold cross-validation. The \textbf{best overall result} is shown in bold, while the \underline{\smash{best result under the same privacy budget}} is underlined. Additional results on Gemma-3-1B and further evaluation details are provided in Appendix~\ref{appendix:more_results}.}
    \vspace{-10pt}
    \label{tab:main_result}
\end{table*}

\section{Experimental Setting}
\label{sec:experimental_setting}
\paragraph{Datasets} 
The MIMIC dataset series is one of the most widely used resources for clinical NLP. We use discharge notes from MIMIC-III \citep{johnson2016mimic} and MIMIC-IV \citep{Johnson2023MIMICIVNote} for different purposes. MIMIC-III serves as the \textit{public} dataset to train the term generation model $\textsc{TermGen}$, while MIMIC-IV is treated as the \textit{private} dataset for training the note generation model $\textsc{NoteGen}$ under DP constraints. For both datasets, we filter out notes without associated ICD codes. We further exclude notes annotated in the SNOMED CT Entity Linking Challenge \citep{Hardman2025SNOMED} from MIMIC-IV notes and reserve them as the test set. This results in three datasets (statistics in Appendix \ref{appendix:dataset_stats}) for our experiments: $D_{\text{public}}$, with 52.7k MIMIC-III notes (500 held out for development); $D_{\text{train}}^{\text{src}}$, with 122k MIMIC-IV notes after exclusion; $D_{\text{test}}^{\text{src}}$, consisting of 204 SNOMED-annotated notes.

\paragraph{Hyperparameters} 
Following common practice in previous work \citep{yu2021do, de2022unlockinghighaccuracydifferentiallyprivate, DBLP:journals/corr/abs-2409-07809}, we experiment with different privacy budgets by varying the overall $\epsilon \in[2, 5, 8]$. Accordingly, the privacy budget for term generation ($\epsilon_t$) or note generation ($\epsilon_n$) is set to one of these values, and the corresponding $\delta$ value is set as $\frac{1}{NlogN}$ where $N$ denotes the size of the private dataset. All experiments are conducted on up to two Nvidia A100 80GB GPUs. More training details, including learning rate, number of epochs, batch size, etc., are provided in Appendix \ref{appendix:hyperparameters}.

\paragraph{Baselines}
We compare our method with existing DP approaches for synthetic text generation: (1) DP-SGD with control codes \citep{DBLP:conf/acl/YueILKMS0LS23}, where a language model is fine-tuned under DP constraints using FastDP and guided by prepended control codes; and (2) AUG-PE \citep{DBLP:conf/icml/Xie0BGYINJZL0Y24}, a private evolution-based method that generates synthetic text without model training. For fair comparison, both baselines are adapted to clinical note generation by prepending ICD codes as control codes, consistent with the conditioning setup in \textsc{Term2Note}. We use $\epsilon_\text{meta}$ to denote the privacy budget for ICD codes and $\epsilon_n$ for note generation.

\paragraph{Evaluation}
\textbf{Automatic evaluation} of synthetic data typically considers fidelity, privacy, and utility. Although privacy is formally guaranteed by DP, we additionally assess personal information leakage and conduct a preliminary empirical privacy analysis (Appendix~\ref{appendix:privacy_eval}), showing encouraging results. \\
\textbf{Fidelity} measures how closely synthetic notes resemble real data in structural, syntactic, and semantic aspects. Structural similarity is evaluated using Kullback--Leibler (KL) divergence over text length distributions. Syntactic similarity is measured by Jaccard similarity and KL divergence over unary and binary clinical term sets, where binary terms denote co-occurring term pairs. Semantic similarity is assessed using MAUVE \citep{DBLP:conf/nips/PillutlaSZTWCH21} with domain-relevant semantic embeddings from BioMistral-7B \citep{labrak-etal-2024-biomistral} to compare global distributional similarity. \\
\textbf{Utility} evaluates the usefulness of synthetic notes for downstream clinical tasks, such as \textbf{ICD coding prediction}; evaluation of additional tasks is presented in the Appendix \ref{appendix:more_results}. ICD-9 and ICD-10 codes are normalised into 20 chapter-level groups (mapping provided in Appendix~\ref{appendix:ICD_grouping}). We compare Train-Real-Test-Real and Train-Synthetic-Test-Real settings using micro/macro F1, AUC, and Precision@$k$. Clinical-Longformer \citep{li2023comparative} is used as the classifier due to its suitability for long clinical documents. Evaluation is conducted on the private test set $D_{\text{test}}^{\text{src}}$ (204 notes), which is split into training and testing subsets. The classifier is trained on ${D_{\text{test-train}}^{\text{src}/\text{syn}}}$ and evaluated on ${D_{\text{test-test}}^{\text{src}}}$. Owing to the limited data size, we use 5-fold cross-validation with an 80:20 train--test split.

\textbf{Human Evaluation} is additionally conducted with three licensed physicians to assess the clinical quality of the generated notes. The evaluation follows a pairwise comparison protocol, where each physician is presented with a randomly selected pair of notes and asked to indicate which one is clinically better. The pairs are sampled from outputs generated by three different models under $\epsilon=8$, as well as from the original (real) notes.
Each physician evaluates a minimum of 100 pairs, resulting in a total of 412 pairwise comparisons across all annotators. Based on these annotations, we estimate pairwise model preferences and infer a global ranking using the Bradley-Terry (BT) \citep{bradley1952rank} model. The BT model also allows us to estimate the probability that one model 
$\mathcal{M}_1$ is preferred over another model $\mathcal{M}_2$, based on the aggregated comparison outcomes.

\section{Results}
\subsection{Main Results}
Table \ref{tab:main_result} presents the main results.

\begin{figure}[t]
    \centering
        \includegraphics[width=1.0\columnwidth]{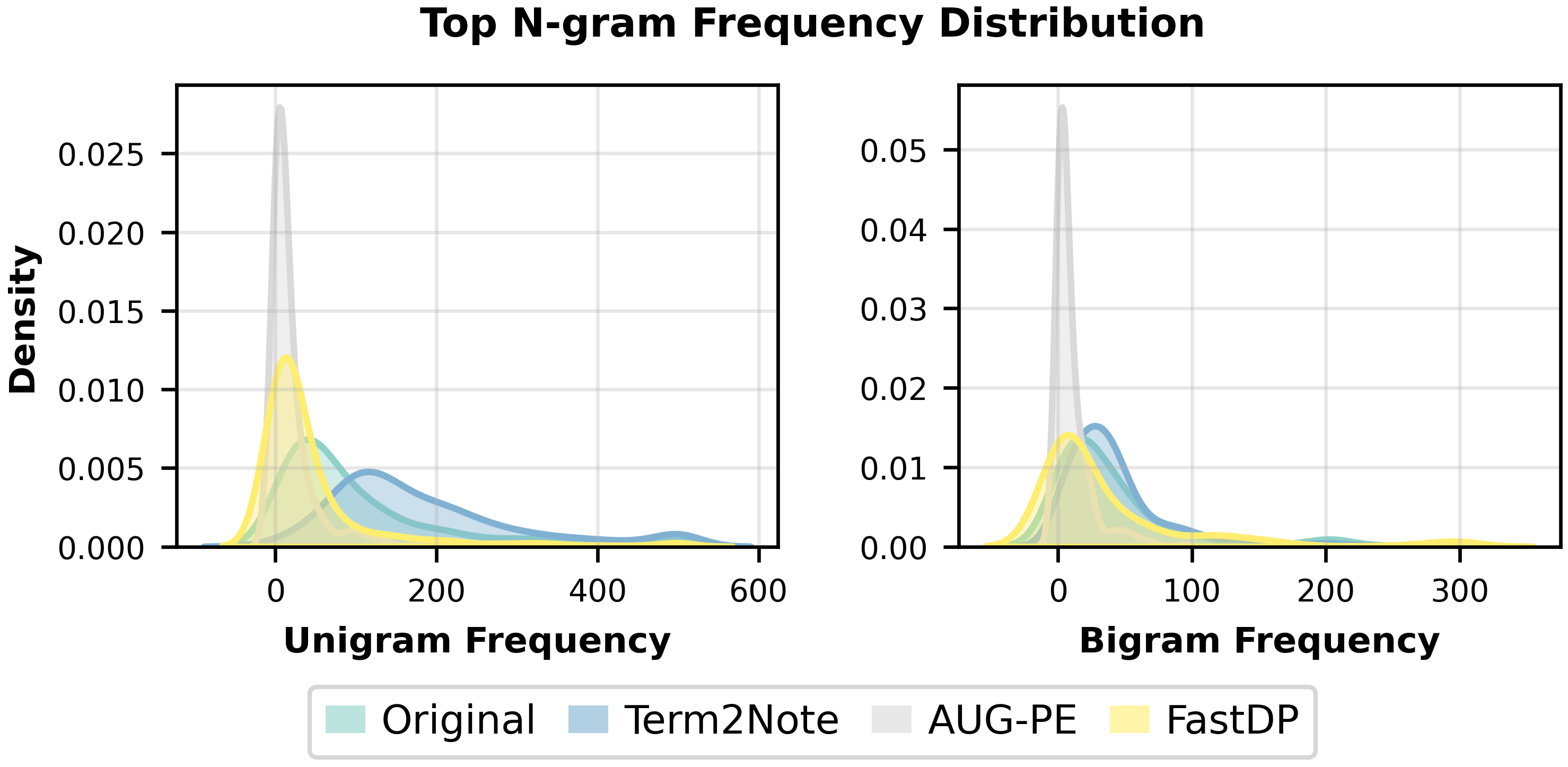}
        \captionof{figure}{Distribution of n-gram frequencies in clinical notes generated by different DP methods under $\epsilon=8$. 
        }
    \label{fig:ngram_freq_dist}
\end{figure}

\paragraph{
Term2Note consistently achieves better structural, syntactic, and semantic similarity to the original data, despite operating under stronger privacy constraints and fewer assumptions.} In terms of \textit{structural similarity}, Term2Note achieves the lowest KL divergence in text length distribution (as low as 0.15), indicating faithful preservation of note structure. For \textit{syntactic similarity}, it obtains the (almost) highest Jaccard scores for both unary and binary clinical terms, alongside the lowest KL divergence in term frequency distribution, suggesting close alignment with real clinical content. In \textit{semantic space}, Term2Note substantially outperforms both baseline methods in MAUVE score, confirming that its generated text is significantly more aligned with the original notes in the semantic space. These advantages remain even under a strict privacy budget of $\epsilon = 2$ and when further enforcing DP on clinical term generation (i.e., $\epsilon_t = 2$), underscoring the robustness of the approach. Additionally, Figure~\ref{fig:ngram_freq_dist} shows the distribution of n-gram frequencies, where Term2Note exhibits a distribution more closely aligned with the original notes compared to the baseline methods. More results are presented in Appendix \ref{appendix:more_results}.

\paragraph{Regarding utility, Term2Note demonstrates strong overall performance and consistently preserves clinical utility across varying privacy budgets, even under stricter constraints on both term and note generation.} Across all privacy levels, Term2Note achieves the highest F1 scores, outperforming both AUG-PE and FastDP, and showing the closest performance to the original data. While AUC and Precision@k scores are comparable, rather than uniformly superior, to those of AUG-PE and FastDP, this variation reflects the different aspects of model behaviour captured by each metric. Notably, even when the privacy budget for note generation is reduced to $\epsilon_n = 2$ and additional constraints are applied to privatise clinical terms (with $\epsilon_t = 2$), Term2Note maintains strong utility, with F1 and AUC scores comparable to or better than the baselines operating under looser privacy conditions. 

\begin{figure}[t]
    \centering
    \includegraphics[width=1.0\columnwidth]{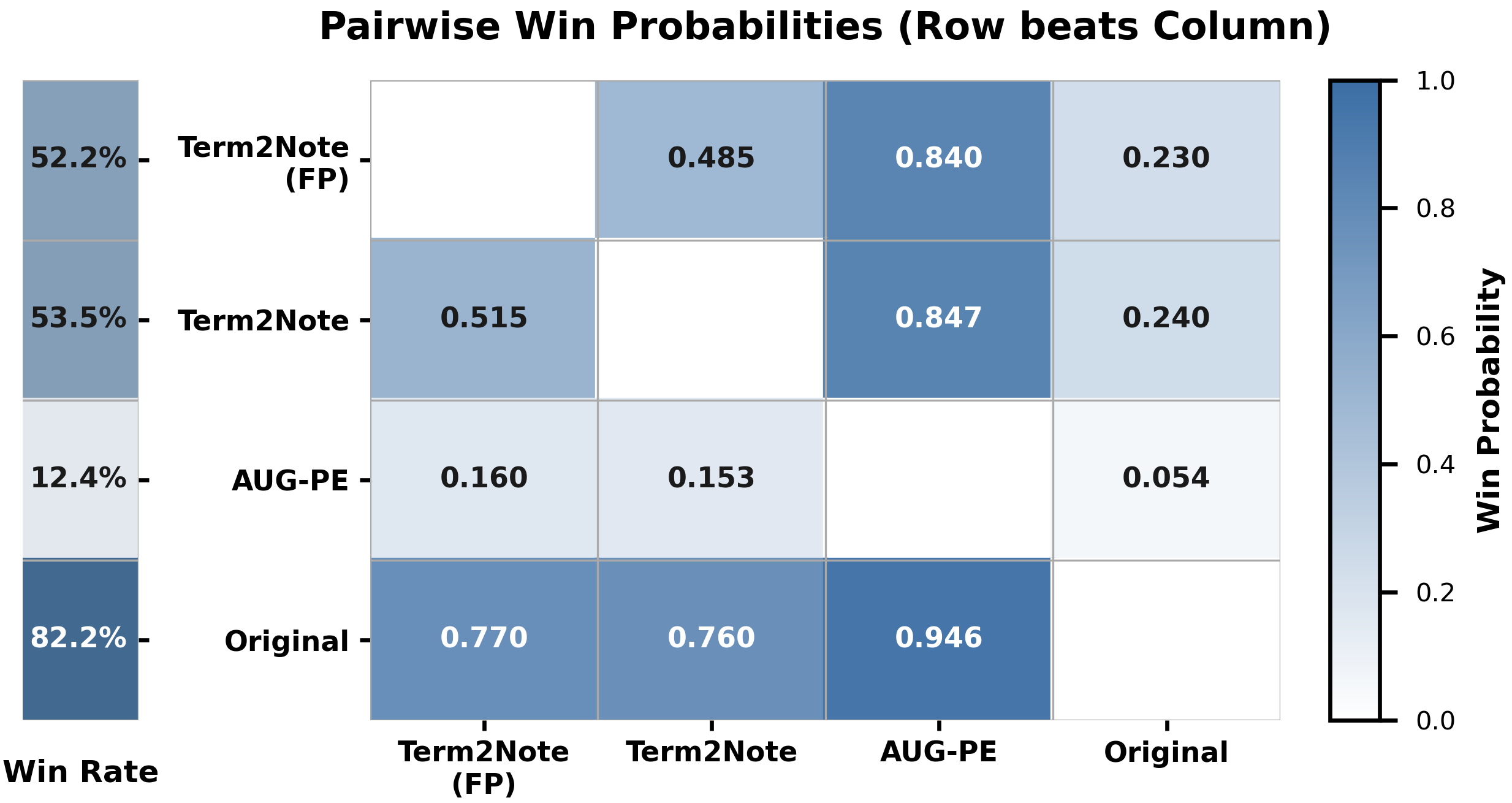}
    \caption{Human evaluation results summarised using the BT model. Term2Note (FP) denotes the \textit{full privacy} setting, where both terms and notes are synthesised under DP constraints.
    }
    \label{fig:human_eval}
\end{figure}

\paragraph{Term2Note is consistently preferred by human experts over AUG-PE, with minimal quality loss under full privacy.} As shown in Figure~\ref{fig:human_eval}, Term2Note achieves a win rate of 52.2\%–53.5\% across DP settings, substantially outperforming AUG-PE (12.4\%). The pairwise win probabilities (right panel) further demonstrate that Term2Note reliably outperforms AUG-PE across all conditions.  Importantly, introducing full privacy constraints, where both clinical terms and notes are protected, has only a marginal effect on human preference. These results suggest that Term2Note maintains high perceived quality while offering stronger privacy guarantees.

\paragraph{Term privatisation maintains a good level of semantic coherence while effectively abstracting away from potentially privacy-leaking details.} We evaluate the semantic alignment of the term generation model \textbf{\textsc{TermGen}} on $D_{\text{test}}^{\text{src}}$ by computing the cosine similarity between the embeddings of original and generated term lists, using the clinical term encoder \textsc{Emb}. Without the DP mechanism $\textsc{DPRP}^*$, the mean cosine similarity is high (0.82), indicating strong recovery of original terms. When $\textsc{DPRP}^*$ is applied, the mean similarity drops to 0.61, reflecting the expected privacy-induced noise. This drop indicates that the generated terms stay semantically coherent without closely matching the originals, reducing the risk of revealing sensitive information (see Appendix~\ref{appendix:term_generation_example} for case studies).

\begin{table}[t]
    \centering
    \resizebox{\columnwidth}{!}{
    \setlength{\tabcolsep}{1mm}
    \small
    \begin{tabular}{l c l cc l c}
        \toprule
        \multirow{3}{*}{\textbf{Method}}
        & \multicolumn{6}{c}{\textbf{Fidelity}} \\
        \cmidrule{2-7}
        & \textbf{Length}
        &
        & \multicolumn{2}{c}{\textbf{Unary/Binary Term}} &
        & \textbf{Semantic} \\
        \cmidrule{2-2}\cmidrule{4-5}\cmidrule{7-7}
        & \textbf{KL Div.}$\downarrow$ &
        & \textbf{Jaccard}$\uparrow$ & \textbf{KL Div.}$\downarrow$ &
        & \textbf{MAUVE}$\uparrow$ \\
        \midrule
        \multicolumn{7}{c}{$\epsilon_{\text{meta}}=\infty, \epsilon_n=8$} \\
        \midrule
        Term2Note
        & 0.39 & & 0.40/0.13 & 0.47/\textbf{1.14} & & \textbf{0.53} \\
        w/o \textsc{SecSplit}
        & 7.10 & & 0.38/0.16 & 2.41/2.05 & & 0.02 \\
        w/o DP Maximiser
        & 0.39 & & 0.41/0.13 & 0.43/\textbf{1.14} & & 0.42 \\
        w/ $\textsc{LLM}_{\text{ppl}}$ & & & & & & \\
        \quad(Meta-Llama-3-8B)
        & \textbf{0.24} & & \textbf{0.47}/\textbf{0.17} & \textbf{0.38}/1.16 & & 0.38 \\
        \midrule
        \multicolumn{7}{c}{$\epsilon_{\text{meta}}=\infty, \epsilon_n=\infty$} \\
        \midrule
        Term2Note
        & 0.25 & & 0.52/0.20 & 0.22/\textbf{1.08} & & \textbf{0.59} \\
        Term2Note & & & & & & \\
        \quad(Llama-3.3-70b)
        & 0.87 & & \textbf{0.55}/\textbf{0.23} & \textbf{0.17}/2.23 & & 0.38 \\
        RAG
        & 0.75 & & 0.43/0.17 & 0.62/1.60 & & 0.22 \\
        \bottomrule
    \end{tabular}
    }
    \caption{Ablation and scaling analysis results.}
    \label{tab:ablation_result}
\end{table}

\subsection{Ablation and Scaling Analysis}

To evaluate the contribution of each component in \textsc{Term2Note}, we conduct an ablation study using Llama-3.2-1B with $\epsilon_{meta}=\infty$ and $\epsilon_n=8$. Fidelity results are reported in Table~\ref{tab:ablation_result}. Removing \textsc{SecSplit}, i.e., generating the full note in a single pass, causes the largest quality drop, reflected by substantially lower MAUVE and higher KL divergence in clinical term distributions, indicating greater deviation from real notes. Removing the DP maximiser or replacing $\textsc{LLM}_{\text{ppl}}$ with a general-domain LLM also degrades performance, though less severely. Overall, the results highlight the importance of all three components in \textsc{Term2Note}.

Table~\ref{tab:ablation_result} also includes a preliminary comparison with Llama-3.3-70B\footnote{\url{https://huggingface.co/meta-llama/Llama-3.3-70B-Instruct}} under a non-private setting ($\epsilon_{meta}=\infty, \epsilon_n=\infty$) to justify our choice of model scale. We evaluate two approaches: Retrieval-Augmented Generation (RAG) and LoRA fine-tuning. For RAG, the top-5 most similar sections from the training set $D^{\text{src}}_{\text{train}}$ are retrieved to assist section-wise generation. For LoRA fine-tuning, the model is quantised to 4-bit precision and trained on 10k samples due to limited computational resources. Although neither approach provides privacy guarantees, they serve as reference points for large-scale models. The results show that LoRA fine-tuning substantially outperforms RAG, demonstrating the importance of parameter adaptation for clinical note generation. The LoRA-finetuned 70B model achieves performance comparable to our fully fine-tuned 1B model and could potentially perform better if trained on the full dataset. However, fine-tuning the 70B model is prohibitively expensive: training on only 10k samples requires more than 80 hours, making large-scale training impractical. In contrast, the 1B model offers a much more practical trade-off between performance and computational cost.

\subsection{Leakage of Private Information}
\begin{table}[t]
    \centering
    \resizebox{\columnwidth}{!}{
    \begin{tabular}{cccc}
        \toprule
        \textbf{Repetition} & $\epsilon$ & \textbf{Perplexity Rank} & \textbf{Leaked Canaries} \\
        \midrule
        \multirow{2}{*}{100} & $\infty$ & 1/10000   & 60\% \\
                             & 8        & 3425/10000  & 0\% \\
        \bottomrule
    \end{tabular}
    }
    \caption{Privacy leakage across privacy budgets.}
    \label{tab:canary}
\end{table}

Following the canary injection experiment of \citet{DBLP:conf/acl/YueILKMS0LS23}, we evaluate potential privacy leakage in synthetic notes generated by Term2Note. We inject canary sequences containing sensitive information (e.g., names, addresses, dates of birth, emails, and unit numbers) into the training set, with each repeated 100 times.
As shown in Table~\ref{tab:canary}, without DP ($\epsilon=\infty$), three out of five canaries are reproduced in the synthetic notes, yielding a leakage rate of 60\%, with an extremely high exposure risk indicated by a perplexity rank of 1/10000. In contrast, under DP ($\epsilon=8$), no canaries are leaked, and the perplexity rank decreases substantially to 3425/10000. These results demonstrate the effective privacy protection of our method.

\subsection{Robustness to the Choice of Public Dataset}
    
\begin{table}[t]
    \centering
    \resizebox{\columnwidth}{!}{
    \begin{tabular}{cc cc l cc l cc}
        \toprule
        \multicolumn{10}{c}{\textbf{Fidelity}} \\
        \midrule
        \multirow{2}{*}{$\mathbf{\epsilon_t}$} & \multirow{2}{*}{$\mathbf{\epsilon_n}$} & \multicolumn{2}{c}{\textbf{Length}} & & \multicolumn{2}{c}{\textbf{Unary/Binary Term}} & & \multicolumn{2}{c}{\textbf{Semantic}} \\
        \cmidrule{3-4}\cmidrule{6-7}\cmidrule{9-10}
         & & \multicolumn{2}{c}{\textbf{KL Div.}$\downarrow$} & & \textbf{Jaccard}$\uparrow$ & \textbf{KL Div.}$\downarrow$ & & \multicolumn{2}{c}{\textbf{MAUVE}$\uparrow$} \\
        \midrule
        $\infty$ & $\infty$ & \multicolumn{2}{c}{1.25} & & 0.47/0.14 & 0.56/3.47 & & \multicolumn{2}{c}{0.37} \\
        $\infty$ & 8 & \multicolumn{2}{c}{0.75} & & 0.45/0.18 & 0.63/2.23 & & \multicolumn{2}{c}{0.44} \\
        8 & 8 & \multicolumn{2}{c}{0.30} & & 0.44/0.16 & 0.62/1.79 & & \multicolumn{2}{c}{0.35} \\
        5 & 8 & \multicolumn{2}{c}{0.36} & & 0.44/0.15 & 0.61/1.69 & & \multicolumn{2}{c}{0.40} \\
        \midrule
        
        \multicolumn{10}{c}{\textbf{Utility}} \\
        \midrule
        \multirow{2}{*}{$\mathbf{\epsilon_t}$} & \multirow{2}{*}{$\mathbf{\epsilon_n}$} & \multicolumn{2}{c}{\textbf{F1}} & & \multicolumn{2}{c}{\textbf{AUC}} & & \multicolumn{2}{c}{\textbf{Precision@$k$}} \\
        \cmidrule{3-4}\cmidrule{6-7}\cmidrule{9-10}
         & & \textbf{Micro} & \textbf{Macro} & &  \textbf{Micro} & \textbf{Macro} & &  $\mathbf{k=3}$ & $\mathbf{k=5}$  \\
        \midrule
        $\infty$ & $\infty$ & 49.17 & 23.04 & & 78.77 & 52.14 & & 66.83 & 58.63 \\
        $\infty$ & 8 & 52.77 & 25.44 & & 79.62 & 53.33 & & 68.47 & 59.24 \\
        8 & 8 & 48.42 & 21.46 & & 78.55 & 49.58 & & 63.08 & 58.23 \\
        5 & 8 & 53.33 & 26.28 & & 78.55 & 50.98 & & 67.67 & 58.24 \\
        \bottomrule
    \end{tabular}
    }
    \caption{Fidelity and utility of Term2Note with \textsc{TermGen} trained on PMC-Patients.}
    \label{tab:pmc-patients}
\end{table}

To evaluate the impact of the public dataset used to train \textsc{TermGen} in \textsc{Term2Note}, we replace MIMIC-III with PMC-Patients \cite{zhao2023large}, a public dataset containing over 250k patient summaries extracted from PMC articles. As PMC-Patients originates from a different source and exhibits a potentially different data distribution, it provides a useful test of the robustness of our framework. We follow the same term extraction procedure, except that each patient summary is treated as a single section, and reserve 500 summaries for development. The extracted medical terms are then used to train \textsc{TermGen}. Results are reported in Table~\ref{tab:pmc-patients}.
Compared with the corresponding settings in Table~\ref{tab:main_result}, where \textsc{TermGen} is trained on MIMIC-III, the performance differences are mixed. For example, PMC-Patients achieves a higher MAUVE score when $\epsilon_t=\infty$ and $\epsilon_n=8$, whereas MIMIC-III performs slightly better under other privacy budgets. Overall, although training \textsc{TermGen} on MIMIC-III generally yields better performance, using PMC-Patients produces comparable fidelity and utility, with only small differences across all evaluation metrics.

\subsection{Qualitative Analysis} 

To further assess the quality of synthetic notes, two physicians each reviewed 20 samples generated by Term2Note, covering both standard and full privacy settings with $\epsilon = 5$. As intended by the design of the evaluation, where \textit{\textbf{physicians were explicitly asked to identify any potential clinical issues, the feedback focuses on shortcomings rather than general plausibility}}. Importantly, not all notes contained identifiable issues. In one physician's review, 45\% were judged to have no clinical problems, indicating that a substantial proportion of generations were considered clinically plausible. \textbf{Both physicians agreed that many synthetic notes were plausible as discharge summaries and generally exhibited sound structural organisation.} However, recurring issues emerged around clinical accuracy and coherence. The first physician highlighted problems such as missing or misordered sections, internal inconsistencies (e.g., conflicting medications), and vague or overly generic phrasing; under full privacy constraints, repetition was more common. The second physician, who reviewed a different set of examples, reported more content-level issues, including medication misclassifications (e.g., labelling omeprazole as an antibiotic), illogical or irrelevant narrative insertions, and errors in clinical reasoning. These observations suggest that while Term2Note performs well in preserving structural fidelity, improvements are needed in clinical fact consistency and terminology use. Figure~\ref{fig:examples} illustrates examples of final note sections, demonstrating the model’s ability to maintain coherence in longer contexts. Although Term2Note occasionally omits sections, its outputs more closely align with the structure of the original note compared to the baseline model.

\section{Conclusions}
In this paper, we introduce Term2Note, a novel framework for DP clinical note generation that synthesises section-wise content conditioned on medical terms while providing formal privacy guarantees.
Experiments show that Term2Note consistently outperforms existing baselines, achieving the highest fidelity and closely matching real notes in structure, semantics, and medical term distributions. Models trained on its synthetic data achieve utility comparable to real data on downstream tasks, while human evaluation shows that clinical experts consistently prefer its outputs over baseline methods.
Overall, Term2Note provides a promising solution to healthcare NLP data scarcity through high-quality, privacy-preserving clinical note generation and facilitates privacy-conscious data sharing.

\section*{Limitations}
Utility metrics exhibit some instability across privacy budgets; performance does not always degrade monotonically as the budget decreases. This may stem from label imbalance relative to training data size, which poses challenges for multi-label prediction when certain labels have limited training examples. Besides, the correlation between clinical factual correctness of the generated synthetic notes and components in Term2Note has not been clearly established due to the limited resources of human experts. Moreover, Term2Note is currently tailored to clinical notes, while other forms of clinical narratives remain unexplored. Nevertheless, it demonstrates the broader applicability of generating clinical content from medical terms, which could be extended to additional clinical text formats in future work.
Qualitative feedback also reveals limitations in clinical reliability. In particular, ensuring factual correctness and avoiding clinical inconsistencies (e.g., incorrect medication classes or contradictory clinical states) remain challenging. Future work could integrate clinical fact-checking mechanisms, structured consistency checks (e.g., validating medications against conditions), or lightweight clinical reasoning modules, either during decoding or as post-generation filters. We view these extensions as natural directions to further improve the clinical safety and usefulness of Term2Note.

\section*{Ethical Considerations}
It is important to note that Term2Note is a research artifact rather than a deployment-ready clinical system. The synthetic clinical notes it generates are intended solely for research purposes and should not be used directly for clinical decision-making without further validation. The formal differential privacy guarantees apply only to the private dataset (i.e., MIMIC-IV in this work) at the note level; they do not provide per-patient privacy guarantees, nor do they apply to the public datasets (i.e., MIMIC-III and PMC-Patients). In addition, \textsc{Term2Note} may generate factual inaccuracies or clinically implausible content. Therefore, rigorous validation, subgroup evaluation, and appropriate human oversight are required before any downstream clinical use. Finally, because the training data are derived from a single healthcare system, the generated notes may reflect demographic or institutional biases. Consequently, evaluation across diverse populations is necessary before broader deployment.

\section*{Acknowledgments}
We thank the anonymous reviewers and the ethics reviewer of the May 2026 cycle for their feedback. This research is part of the IN-CYPHER programme and is supported by the National Research Foundation, Prime Minister’s Office, Singapore under its Campus for Research Excellence and Technological Enterprise (CREATE) programme. This research is also supported by the North West Cyber Security Connect for Commercialisation (NW CyberCom), funded by the Research England Connecting Capability Fund, and UKRI Digital Research Infrastructure Fund, through the FORTRESS project under the DARE UK next-gen catalyst programme. We are grateful for the support provided by Research IT in form of access to the Computational Shared Facility at The University of Manchester and the computational facilities at the Imperial College Research Computing Service\footnote{DOI: \url{https://doi.org/10.14469/hpc/2232}}.


\bibliography{references}

\appendix

\section{Example Demonstration}
\label{appendix:example_demo}
\begin{figure*}
    \centering
    \includegraphics[width=0.98\linewidth]{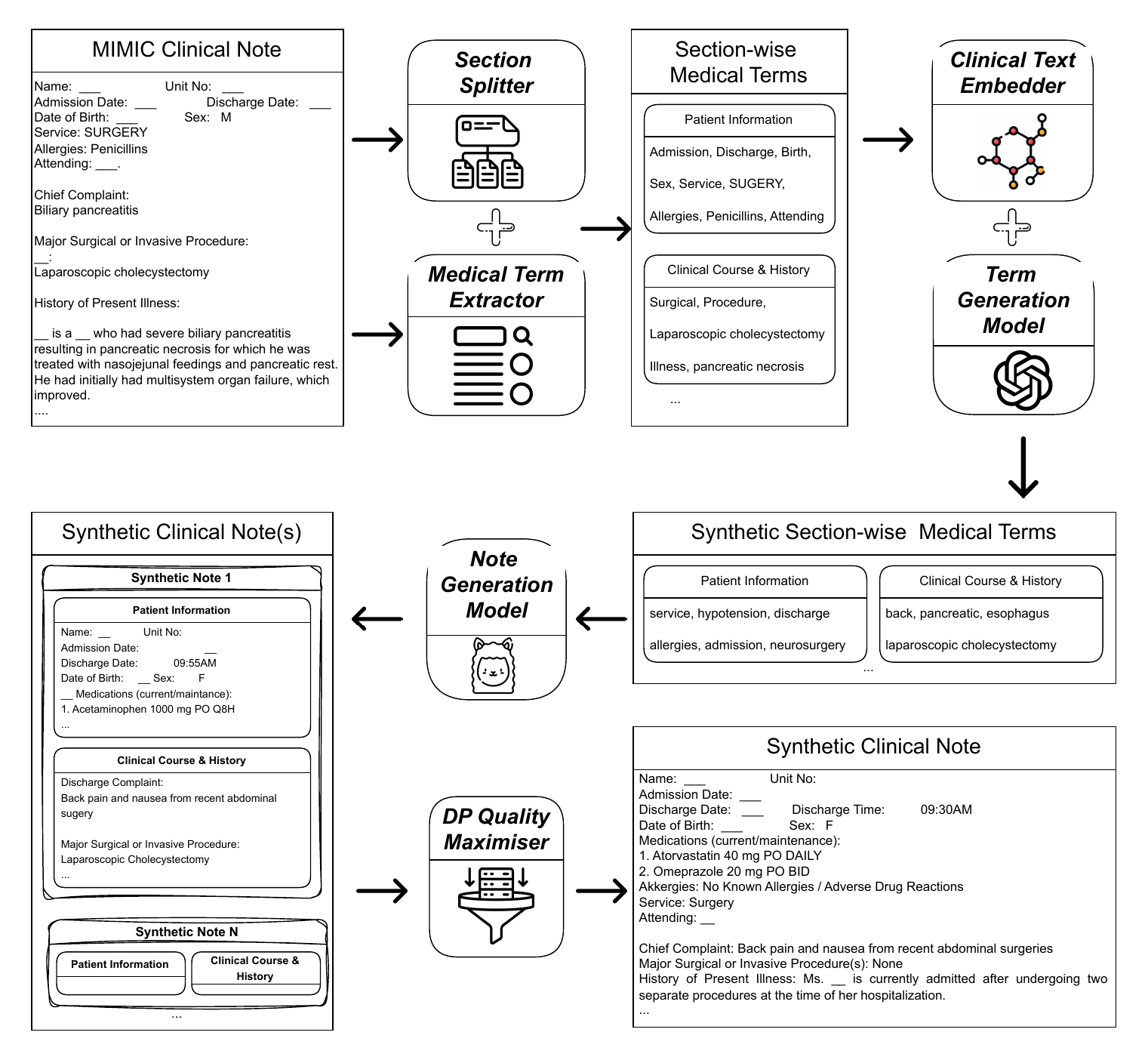}
    \caption{Example to demonstrate Term2Note.}
    \label{fig:example_demo}
\end{figure*}

Figure \ref{fig:example_demo} presents the detailed process in Term2Note with example for each step.

\begin{figure*}
    \centering
    \includegraphics[width=1.0\linewidth]{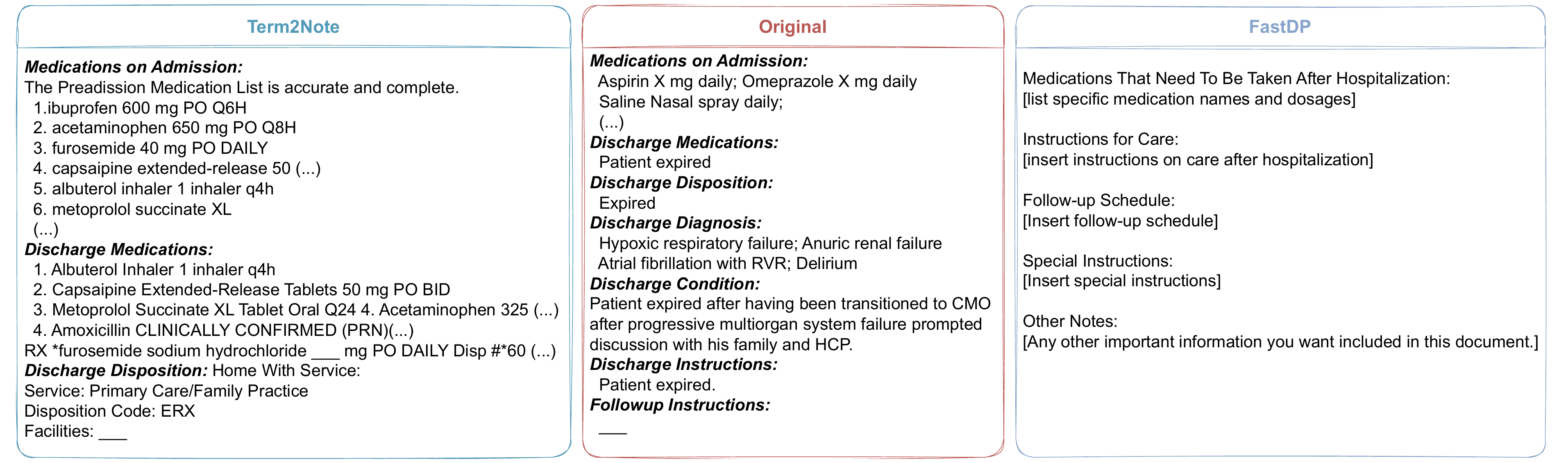}
    \caption{Examples of the last section in clinical notes generated by different models. For illustration purposes, some content is redacted with ``(...)'', and numeric values in the original note are de-identified. No other modifications were made.}
    \label{fig:examples}
\end{figure*}

Figure \ref{fig:examples} presents examples of generation from different models.

\section{Algorithm Pseudocode}
Algorithm \ref{algo:Term2Note} presents the pseudocode for Term2Note.
\begin{algorithm*}[htbp]
    \caption{Term2Note}
    \label{algo:Term2Note}
    
    \textbf{Input:} Private note $X^{\text{src}}$, public term lists $D_\text{public}$, privacy parameters for term generation $(\epsilon_t, \delta_t)$ and note generation $(\epsilon_n, \delta_n)$, \#sections $m$, instruction $I$, noise variance $\sigma_{\text{emb}}$, \#candidates $k$
    
    \textbf{Output:} Synthetic note $X^{\text{syn}}$ with $(\epsilon, \delta)$-DP guarantee
    
    \begin{algorithmic}[1]

        \STATE $\text{SEC}_{1:m}^{\text{src}} \gets \textsc{SecSplit}(X^{\text{src}})$ \hfill \textbf{\small // \ref{sec:format_and_term} Format and Term Identification}
        \STATE $T_{1:m}^{\text{src}} \gets \textsc{TermExt}(\text{SEC}^{\text{src}}_{1:m})$

        \STATE $E_{1:m}^{\text{src}}, E_{\text{public}} \gets \textsc{Emb}(T_{1:m}^{\text{src}}), \textsc{Emb}(D_{\text{public}})$ \hfill \textbf{\small // \ref{sec:term_gen} Clinical Terms Generation}
        \IF{train}
            \STATE $\theta_t, \theta_p \gets \mathcal{L}(\textsc{TermGen}_{\theta_t}(\textsc{Proj}_{\theta_p}(E_{\text{public}}+\mathcal{N}(0, \sigma_{\text{emb}}))), D_{\text{public}})$ \hfill \small // Train on $D_{\text{public}}$
        \ELSE
            \STATE $T_{1:m}^{\text{syn}} \sim \textsc{TermGen}_{\theta_t}(\textsc{Proj}_{\theta_p}(\textsc{DPRP}^*(E_{1:m}^{\text{src}}, \frac{\epsilon_t}{m}, \frac{\delta_t}{m})))$ \hfill \small // Infer for $X^{\text{src}}$
        \ENDIF

        \IF{train}
            \STATE $\theta_n \gets \text{FastDP}(\mathcal{L}(\textsc{NoteGen}_{\theta_n}(I, \text{SEC}_{<i}^{\text{src}}, T_{i}^{\text{src}}), X^{\text{src}}), \epsilon_n, \delta_n)$ \hfill \textbf{\small // \ref{sec:note_gen} Clinical Note Generation}
        \ENDIF
        \STATE $\text{SEC}_j^{\text{syn}}[i] \sim \textsc{NoteGen}_{\theta_n}(I, \text{SEC}_{<j}^{\text{syn}}[i], T_{j}^{\text{src / syn}})$ \textbf{ for } $i = 1, \ldots, k$; $j = 1, \ldots, m$
        \STATE $X^{\text{syn}} \gets \arg \min_{i\in[k]} \textsc{PPL}(\text{SEC}_{1:m}^{\text{syn}}[i])$ \hfill \textbf{\small // \ref{sec:dp_ranker} DP Quality Maximiser}
        
        \RETURN $X^{\text{syn}}$
    
    \end{algorithmic}
\end{algorithm*}

\label{appendix:dprp_algo}
Algorithm \ref{algo:dprp} presents the pseudocode for $\textsc{DPRP}^*$.
\begin{algorithm*}[h]
    \caption{$\textsc{DPRP}^*$}
    \label{algo:dprp}
    \textbf{Input:} Embeddings $E$, privacy parameters $(\epsilon, \delta)$ , privacy allocation $b=0.85$
    
    \textbf{Output:} Privatised Embeddings $E_{\text{DP}}$ with $(\epsilon, \delta)$-DP
    
    \begin{algorithmic}[1]

        \STATE $(\epsilon_1, \delta_1), (\epsilon_2, \delta_2) \gets 0.85 * (\epsilon, \delta), 0.15 * (\epsilon, \delta)$
        \STATE Derive $\sigma_i$ from $(\epsilon_{1i}, \delta_{1i})$; $i \in [1, 2]$
        \STATE $E' = E + \mathcal{N}(0, \sigma_1^2)$
        \STATE $E_C' = E^TE + \mathcal{N}(0, \sigma_2^2)$
        \STATE $V'\Sigma'{V'}^T = \text{SVD}(E_C')$
        \STATE $V'_{k} = V'[1,...,k]; k=0.6*E_\text{hdim}$
        \STATE $E_{\text{DP}} = E'{V'_k}^{T+}{V'_k}^T$  \hfill // + refers to the Moore-Penrose pseudoinverse

        \RETURN $E_{\text{DP}}$
    
    \end{algorithmic}
\end{algorithm*}

\section{Privacy Proof}
\label{appendix:privacy_proof}

\begin{proof}
    We prove that for the full privatisation setting ($T_i = T_i^{\text{syn}}$), Term2Note achieves $(\max(\epsilon_n, \epsilon_t), \max(\delta_n, \delta_t))$-DP by applying the parallel composition theorem.
    
    \textbf{Step 1: Parallel Composition Lemma}
    
    First, we establish the parallel composition property.
    
    \begin{lemma}[Parallel Composition]
        Let dataset $D = D_1 \cup D_2$, and $D_1 \cap D_2 = \emptyset$. Let $\mathcal{M}_1: \mathbb{N}^{|\mathcal{X}_1|} \to R_1$ be $(\epsilon_1, \delta_1)$-DP and $\mathcal{M}_2: \mathbb{N}^{|\mathcal{X}_2|} \to R_2$ be $(\epsilon_2, \delta_2)$-DP. Then $\mathcal{M}(D) = (\mathcal{M}_1(D_1), \mathcal{M}_2(D_2))$ is $(\max(\epsilon_1, \epsilon_2), \max(\delta_1, \delta_2))$-DP.
    \end{lemma}
    
    \begin{proof}[Proof of Lemma]
    Let $D$ and $D'$ be neighboring datasets differing by one record. Since $D_1 \cap D_2 = \emptyset$, the differing record is in either $D_1$ or $D_2$, but not both.
    
    \textbf{Case 1:} The differing record is in $D_1$, so $D_1 \neq D'_1$ but $D_2 = D'_2$.
    
    For any measurable sets $B_1 \subseteq R_1, B_2 \subseteq R_2$:
    \begin{align*}
        &P[\mathcal{M}(D) \in B_1 \times B_2] \\
        &= P[\mathcal{M}_1(D_1) \in B_1] \cdot P[\mathcal{M}_2(D_2) \in B_2] \quad \\
        &\leq \left(e^{\epsilon_1} P[\mathcal{M}_1(D'_1) \in B_1] + \delta_1\right) \cdot P[\mathcal{M}_2(D_2) \in B_2] \quad \\
        &= \left(e^{\epsilon_1} P[\mathcal{M}_1(D'_1) \in B_1] + \delta_1\right) \cdot P[\mathcal{M}_2(D'_2) \in B_2] \quad \\
        &= e^{\epsilon_1} P[\mathcal{M}(D') \in B_1 \times B_2] + \delta_1 P[\mathcal{M}_2(D'_2) \in B_2] \\
        &\leq e^{\epsilon_1} P[\mathcal{M}(D') \in B_1 \times B_2] + \delta_1 \quad 
    \end{align*}

    \textbf{Case 2:} The differing record is in $D_2$, so $D_1 = D'_1$ but $D_2 \neq D'_2$.
    Similarly:
    \begin{align*}
        & P[\mathcal{M}(D) \in B_1 \times B_2] \\
        & \leq e^{\epsilon_2} P[\mathcal{M}(D') \in B_1 \times B_2] + \delta_2 \\
    \end{align*}
    
    \textbf{Combining cases:} For arbitrary neighbouring datasets, we have:
    \begin{align*}
        & P[\mathcal{M}(D) \in B_1 \times B_2] \\
        & \leq e^{\max(\epsilon_1, \epsilon_2)} P[\mathcal{M}(D') \in B_1 \times B_2] + \max(\delta_1, \delta_2) \\
    \end{align*}
    
    Therefore, $\mathcal{M}$ is $(\max(\epsilon_1, \epsilon_2), \max(\delta_1, \delta_2))$-DP.
    \end{proof}
    
    \textbf{Step 2: Application to Term2Note}
    
    Now we apply the parallel composition lemma to Term2Note.
    
    We have:
    \begin{itemize}
        \item $\mathcal{M}_1 = $ \textsc{NoteGen} training on $D_{\text{train}}$, which is $(\epsilon_n, \delta_n)$-DP
        \item $\mathcal{M}_2 = $ \textsc{TermGen} processing on $D_{\text{test}}$, which is $(\epsilon_t, \delta_t)$-DP
        \item $D_{\text{train}} \cap D_{\text{test}} = \emptyset$
    \end{itemize}
    
    Term2Note can be written as:
    $$\text{Term2Note}(D) = f(\mathcal{M}_1(D_{\text{train}}), \mathcal{M}_2(D_{\text{test}}))$$
    
    where $f$ is a deterministic function that applies the trained \textsc{NoteGen} model to the synthetic terms from \textsc{TermGen}.
    
    Since $f$ is a post-processing function applied to the outputs of the parallel composition, and post-processing preserves differential privacy, we have:
    $$\text{Term2Note}(D) \text{ is } (\max(\epsilon_n, \epsilon_t), \max(\delta_n, \delta_t))\text{-DP}$$
\end{proof}

\section{Section Grouping}
\label{appendix:section_grouping}
Table \ref{tab:sec_group} presents the section grouping taxonomy for our \textsc{SecSplit}.

\begin{table}[h]
\centering
\resizebox{\columnwidth}{!}{%
    \begin{tabular}{|p{0.25\columnwidth}|p{0.7\columnwidth}|}
        \toprule
        \textbf{Group Name} & \textbf{Sections} \\
        \midrule
        Patient Information & ``Name'', ``Unit No'', ``Admission Date'', ``Discharge Date'', ``Date of Birth'', ``Sex'', ``Service'', ``Allergies'', ``Attending'' \\
        \midrule
        Clinical Course \& History & ``Chief Complaint'', ``Major Surgical or Invasive Procedure'', ``History of Present Illness'', ``Review of Systems'', ``Past Medical History'', ``Social History'', ``Family History'' \\
        \midrule
        Examinations \& Findings & ``Physical Exam'' \\
        \midrule
        Laboratory \& Imaging Results & ``Pertinent Results'' \\
        \midrule
        Hospital Stay \& Treatment & ``Brief Hospital Course'' \\
        \midrule
        Medications \& Discharge Plan & ``Medications on Admission'', ``Discharge Medications'', 
        ``Discharge Disposition, ``Discharge Diagnosis'', ``Discharge Condition'', 
        ``Discharge Instructions'', ``Followup Instructions'' \\
        \bottomrule
    \end{tabular}
    }
    \caption{The grouped section titles.}
    \label{tab:sec_group}
\end{table}

\section{Dataset Statistics}
\label{appendix:dataset_stats}
Table~\ref{tab:dataset_stats} summarises key statistics of the three datasets used in this study, including the number of clinical notes, and average note length, among other relevant attributes.

\begin{table}[h]
    \centering
    \resizebox{\columnwidth}{!}{%
        \begin{tabular}{cccc}
            \toprule
            \textbf{Corpus} & \textbf{MIMIC-III} & \multicolumn{2}{c}{\textbf{MIMIC-IV}} \\
            \cmidrule{3-4}
            \textbf{Dataset} & $D_{\text{public}}$ & $D_{\text{train}}^{\text{src}}$ & $D_{\text{test}}^{\text{src}}$ \\
            \midrule
            \textbf{\# notes} & 52,722 & 122,202 & 204 \\
            \textbf{avg. \# tokens} & 3327.93 & 3360.60 & 2818.63  \\
            \textbf{avg. \# sections} & 4.59 & 5.77 & 5.79 \\
            \textbf{avg. \# terms} & 176.26 & 203.20 & 173.33 \\
            \textbf{avg. \# ICD codes} & - & - & 6.73 \\
            \bottomrule
        \end{tabular}
    }
    \caption{Dataset statistics. avg. refers to the average of. \# tokens is calculated by taking the average of tokens in each note, tokenised by Llama-3.2-1B-Instruct.}
    \label{tab:dataset_stats}
\end{table}

\section{Hyperparameters}
\label{appendix:hyperparameters}
\paragraph{\textsc{TermGen}}
We fine-tune GPT-2-large on section-wise clinical terms extracted from $D_{\text{public}}$ for up to 5 epochs. The final model is selected based on the checkpoint with the highest F1 score, evaluated on a held-out set of 500 notes. During training, we set the embedding perturbation scale $\sigma_{\text{emb}} = 0.05$, with a batch size of 8 and a learning rate of 2e-5. At inference, we use a batch size of 16 and a maximum generation length of 512 tokens. To ensure reproducibility, decoding is performed with a temperature of 0.1 and top-$p$ set to 1.0.

\paragraph{\textsc{NoteGen}}
\textbf{Training:} We fine-tune Llama-3.2-1B-Instruct or Gemma-3-1B-IT on $D_{\text{train}}^{\text{src}}$ for up to 2 epochs using 2 GPUs. The batch size per device is 2, with a gradient accumulation step of 64 and a learning rate of 5e-5. We enable DeepSpeed ZeRO Stage 3 to optimise memory usage.
\textbf{Inference:} We adapt vLLM for faster generation, with decoding parameters set as temperature = 0.1, top-$p$ = 1.0, repetition penalty = 1.2 and max tokens per section = 2048 across all experiments. Llama-3.2 tends to generate overly long outputs during section-wise generation, so we apply a logit bias on the EOS token to encourage early stopping. This bias is set between 0.5 and 6.0: for DP-enabled models, the value is 0.5 or 1.0; for the non-private setting ($\epsilon = \infty$), it is set to 6.0. Additionally, DP-enabled models use a frequency penalty of 0.4 to further discourage repetition.

For the FastDP baseline, which produces relatively short outputs, we apply only a repetition penalty during inference. Before applying the DP quality maximiser, we generate multiple candidates per input: 4 for Term2Note and FastDP, and 7 for AUG-PE, using the same decoding settings described above.

\section{Distance-based Privacy Evaluation}
\label{appendix:privacy_eval}
\begin{figure*}[h]
    \centering
    \includegraphics[width=1.0\textwidth]{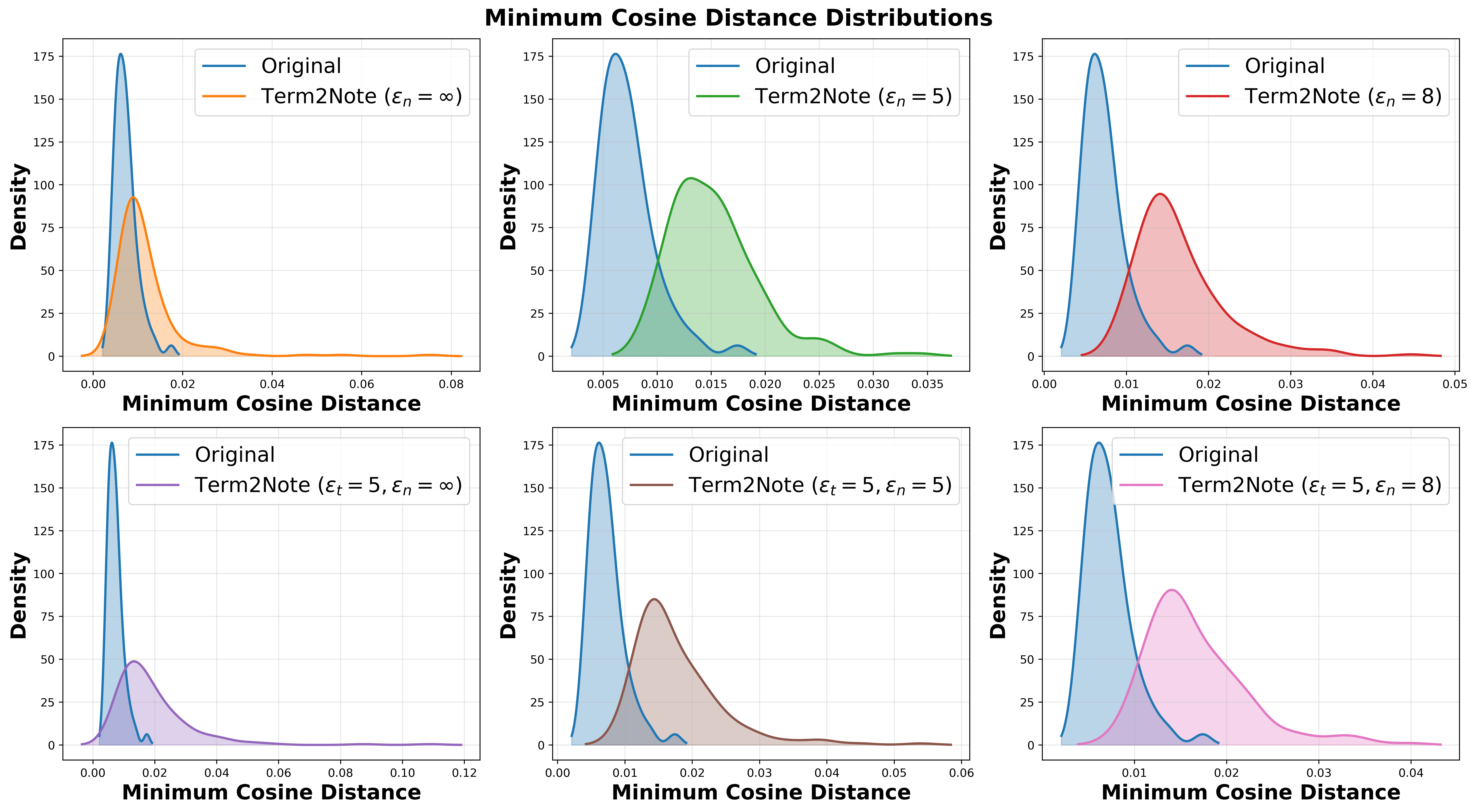}
    \caption{Distribution of minimum cosine distances for all evaluated synthetic strategies compared to the baseline of original test notes. Here, $\epsilon_t$ and $\epsilon_n$ are the privacy budgets for \textsc{TermGen} and \textsc{NoteGen}, respectively.}
    \label{fig:privacy_distributions}
\end{figure*}

A preliminary privacy evaluation is conducted to assess the privacy-preserving properties of the synthetic clinical notes using a membership inference attack (MIA) framework. 
We compare the distribution of minimum cosine distances between the synthetic notes generated by Term2Note and the original training data to a baseline of real, non-member test notes. As illustrated in Figure~\ref{fig:privacy_distributions}, the synthetic notes are, on average, located significantly further from the training data than the test notes. Notably, the setting with $\epsilon_n=\infty$, which is generated without DP, exhibited the most overlap with the test set's distribution. This expected outcome highlights the privacy benefits of the other DP settings, and provides a clear baseline for comparison. 

In future work, we intend to expand our privacy evaluation using canary-based membership inference attacks. This approach involves injecting specially crafted canaries into the training data to establish a worst-case lower bound on privacy risks.

\section{ICD Codes Grouping}
\label{appendix:ICD_grouping}
Table~\ref{tab:icd_group} shows the mapping between our combined ICD categories and the corresponding ICD-9 and ICD-10 chapter headings. For fine-tuning Clinical-Longformer on this classification task, we train for 30 epochs per setting (i.e., model and data fold), with a batch size of 8 and a learning rate of 2e-5.

\begin{table*}[h]
    \centering
    \resizebox{\textwidth}{!}{
    \scriptsize
    \begin{tabular}{|p{0.3\textwidth}|p{0.3\textwidth}|p{0.3\textwidth}|}
        \toprule
        \textbf{Combined ICD Category} & \textbf{ICD-9} & \textbf{ICD-10} \\
        \midrule
        Certain Infectious And Parasitic Diseases & Infectious And Parasitic Diseases & Certain Infectious And Parasitic Diseases \\
        \midrule
        Neoplasms & Neoplasms & Neoplasms \\
        \midrule
        Endocrine, Nutritional And Metabolic Diseases, And Immunity Disorders & Endocrine, Nutritional And Metabolic Diseases, And Immunity Disorders & Endocrine, Nutritional And Metabolic Diseases \\
        \midrule
        Diseases Of The Blood And Blood-Forming Organs And Certain Disorders Involving The Immune Mechanism & Diseases Of The Blood And Blood-Forming Organs & Diseases Of The Blood And Blood-Forming Organs And Certain Disorders Involving The Immune Mechanism \\
        \midrule
        Mental And Behavioural Disorders & Mental Disorders & Mental And Behavioural Disorders \\
        \midrule
        Diseases Of The Nervous System And Sense Organs & Diseases Of The Nervous System And Sense Organs & Diseases Of The Nervous System \\
        \midrule
        Diseases Of The Circulatory System & Diseases Of The Circulatory System &  Diseases Of The Circulatory System \\
        \midrule
        Diseases Of The Respiratory System & Diseases Of The Respiratory System & Diseases Of The Respiratory System \\ 
        \midrule
        Diseases Of The Digestive System & Diseases Of The Digestive System & Diseases Of The Digestive System \\ 
        \midrule
        Diseases Of The Genitourinary System & Diseases Of The Genitourinary System & Diseases Of The Genitourinary System \\
        \midrule
        Complications Of Pregnancy, Childbirth, And The Puerperium & Complications Of Pregnancy, Childbirth, And The Puerperium & Complications Of Pregnancy, Childbirth, And The Puerperium \\ 
        \midrule
        Diseases Of The Skin And Subcutaneous Tissue & Diseases Of The Skin And Subcutaneous Tissue & Diseases Of The Skin And Subcutaneous Tissue \\ 
        \midrule
        Diseases Of The Musculoskeletal System And Connective Tissue & Diseases Of The Musculoskeletal System And Connective Tissue & Diseases Of The Musculoskeletal System And Connective Tissue \\
        \midrule
        Congenital Malformations, Deformations And Chromosomal Abnormalities & Congenital Anomalies & Diseases Of The Musculoskeletal System And Connective Tissue \\ 
        \midrule
        Congenital Malformations, Deformations And Chromosomal Abnormalities & - & Congenital Malformations, Deformations And Chromosomal Abnormalities \\
        \midrule
        Certain Conditions Originating In The Perinatal Period & Certain Conditions Originating In The Perinatal Period & Certain Conditions Originating In The Perinatal Period  \\ 
        \midrule
        Symptoms, Signs And Abnormal Clinical And Laboratory Findings, Not Elsewhere Classified & Symptoms, Signs, And Ill-Defined Conditions & Symptoms, Signs And Abnormal Clinical And Laboratory Findings, Not Elsewhere Classified \\ 
        \midrule
        Injury, Poisoning And Certain Other Consequences Of External Causes & Injury And Poisoning & Injury, Poisoning And Certain Other Consequences Of External Causes \\ 
        \midrule
        External Causes Of Morbidity And Mortality, Injusy and Poisoning & External Causes Of Injury And Poisoning & External Causes Of Morbidity And Mortality, Injusy and Poisoning \\ 
        \midrule
        Factors Influencing Health Status And Contact With Health Services & Factors Influencing Health Status And Contact With Health Services & Factors Influencing Health Status And Contact With Health Services \\
        \midrule
        Diseases Of The Eye And Adnexa & - & Diseases Of The Eye And Adnexa \\
        \midrule
        Diseases Of The Ear And Mastoid Process & - & Diseases Of The Ear And Mastoid Process \\
        \midrule
        Codes For Special Purposes & - & Codes For Special Purposes \\
        \bottomrule
    \end{tabular}
    }
    \caption{The grouped ICD codes.}
    \label{tab:icd_group}
\end{table*}

\section{Privacy Accounting}
\label{appendix:privacy_accounting}
Table \ref{tab:privacy_accounting} presents the summary of how privacy budgets are allocated across components.

\begin{table}[h]
\centering
\resizebox{\columnwidth}{!}{%
\begin{tabular}{ccccc}
\toprule
Term Generator & Note Generator & $\epsilon_n$ & $\epsilon_t$ & $\epsilon$ (total) \\
\midrule
$\times$      & $\checkmark$ & --                      & $\infty$        & $\infty$ \\
$\checkmark$  & $\checkmark$ & $\infty / 8 / 5 / 2$    & $\infty$        & $\infty$ \\
$\checkmark$  & $\checkmark$ & $8$                     & $8 / 5 / 2$     & $8$      \\
$\checkmark$  & $\checkmark$ & $5$                     & $5 / 2$         & $5$      \\
$\checkmark$  & $\checkmark$ & $2$                     & $2$             & $2$      \\
\bottomrule
\end{tabular}%
}
\caption{Summary of privacy accounting.}
\label{tab:privacy_accounting}
\end{table}

\section{Supplementary Experimental Results}
\label{appendix:more_results}

\subsection{Additional Downstream Tasks}

To further evaluate the utility of the synthetic notes, we conduct experiments on two additional downstream tasks.

\paragraph{Continued pre-training.}
We perform a small-scale continued pre-training experiment using \texttt{dmis-lab/biobert-base-cased-v1.2} on the synthetic notes generated by Term2Note. The resulting model (``BioBERT-tuned'') is evaluated on three biomedical benchmarks: named entity recognition (BC5CDR), relation extraction (GAD), and question answering (BioASQ 7b), and compared with the original BioBERT. Tables~\ref{tab:ner-rl} and~\ref{tab:qa} show that continued pre-training consistently improves F1 on both NER and relation extraction, demonstrating that the synthetic notes provide useful domain-specific supervision.

\begin{table}[h]
\centering
\resizebox{\columnwidth}{!}{%
\begin{tabular}{llccc}
\toprule
Task-Dataset & Model & Precision & Recall & F1 \\
\midrule
\multirow{2}{*}{NER-BC5CDR} & BioBERT       & 85.23 & 88.26 & 86.67 \\
                             & BioBERT-tuned & 85.78 & 88.24 & 86.98 \\
\midrule
\multirow{2}{*}{RL-GAD}     & BioBERT       & 80.00 & 79.23 & 79.34 \\
                             & BioBERT-tuned & 82.13 & 80.81 & 80.95 \\
\bottomrule
\end{tabular}%
}
\caption{NER and relation extraction results.}
\label{tab:ner-rl}
\end{table}

\begin{table}[h]
\centering
\resizebox{\columnwidth}{!}{%
\begin{tabular}{llccc}
\toprule
Task-Dataset & Model & Strict Accuracy & Lenient Accuracy & Mean Reciprocal Rank \\
\midrule
\multirow{2}{*}{QA-BioASQ 7b} & BioBERT       & 65.55 & 97.21 & 78.99 \\
                               & BioBERT-tuned & 62.94 & 97.21 & 77.21 \\
\bottomrule
\end{tabular}%
}
\caption{Question answering results.}
\label{tab:qa}
\end{table}

\paragraph{Summarisation.}
We further evaluate the synthetic notes on the BioNLP 2024 Shared Task 2 (Discharge Me!) \cite{PhysioNet-discharge-me-1.3}. Following the shared-task setup, the ``Brief Hospital Course'' (BHC) section is treated as the reference summary and removed from each clinical note. Term2Note is trained on the remaining content to generate synthetic notes. Owing to the large training set (over 36k notes) and limited computational resources, the DP quality maximiser is omitted in this experiment. We fine-tune \texttt{google/long-t5-tglobal-large} \cite{guo2021longt5} on either the original or synthetic training set and evaluate on the original test set. As shown in Table~\ref{tab:utility_bionlp24}, models trained on synthetic notes perform worse than those trained on the original data across all metrics. However, the performance gaps remain relatively small on BLEU, ROUGE-2, and BERTScore, suggesting that the synthetic notes preserve much of the information required for summarisation.

\begin{table}[t]
    \centering
    \resizebox{\columnwidth}{!}{
    \begin{tabular}{lcccc}
        \toprule
        \textbf{Method} & \textbf{BLEU} & \textbf{ROUGE-1/2/L} & \textbf{BERTScore} & \textbf{Meteor} \\
        \midrule
        Original Data & 6.51 & 27.28/7.32/15.48 & 50.85 & 18.83 \\
        Term2Note ($\epsilon_n=\infty$) & 6.39 & 13.06/5.35/9.19 & 46.09 & 10.56 \\
        Term2Note ($\epsilon_n=8$) & 4.93 & 15.72/5.15/10.74 & 48.03 & 11.36 \\
        \bottomrule
    \end{tabular}
    }
    \caption{Utility evaluation on BioNLP 2024 Shared Task 2 Discharge Me!.}
    \label{tab:utility_bionlp24}
\end{table}

\subsection{Fidelity}

\begin{figure}
    \centering
    \includegraphics[width=\linewidth]{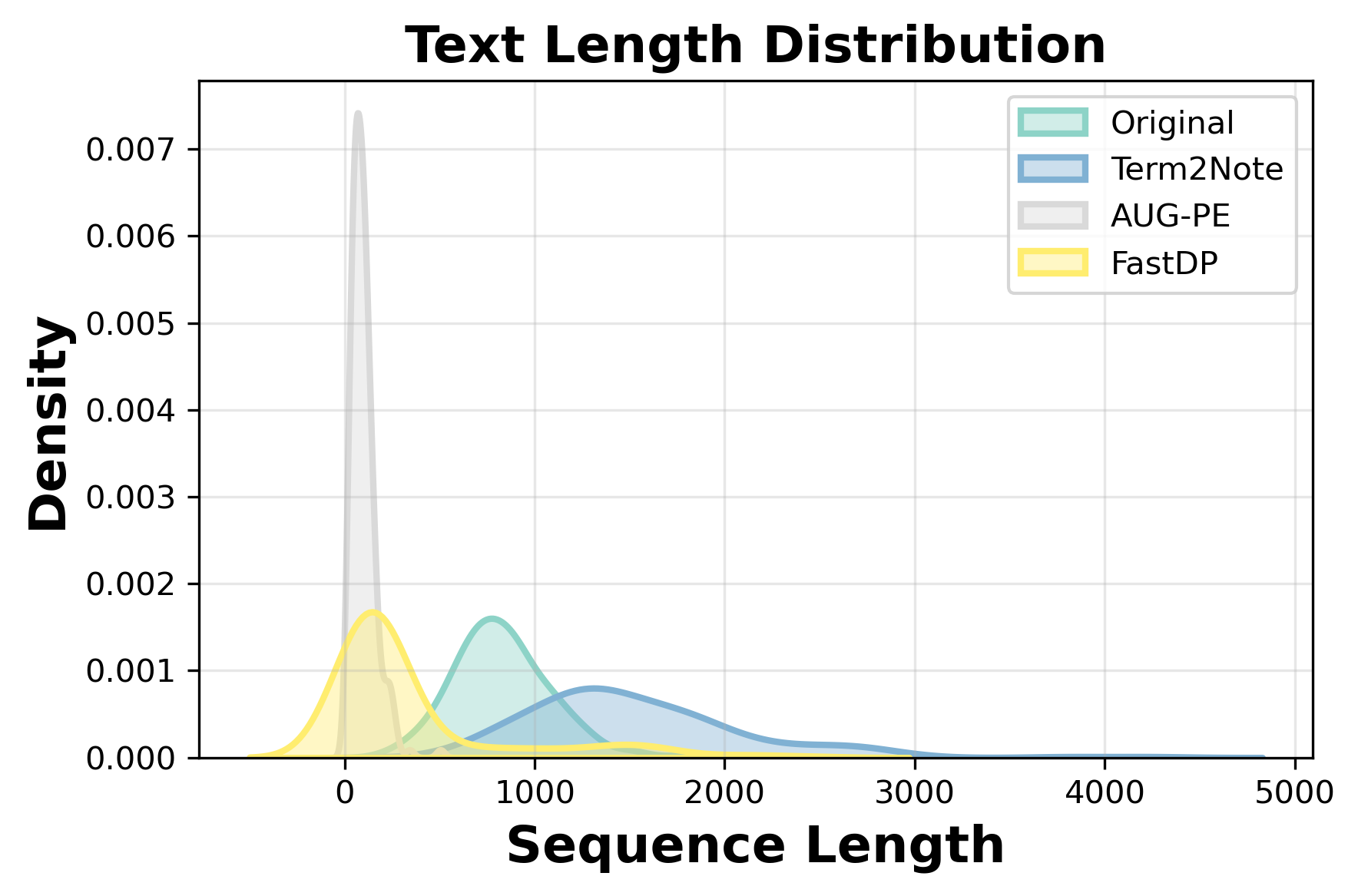}
    \caption{Text length distribution.}
    \label{fig:text_len_dist}
\end{figure}

\paragraph{Text Length Comparison under $\epsilon=8$}
Figure~\ref{fig:text_len_dist} shows the distribution of sequence lengths for clinical notes generated by different methods under a fixed privacy budget of $\epsilon=8$. While all synthetic methods shift the length distribution away from the original data to some extent, Term2Note exhibits the closest alignment. Its distribution more faithfully captures the broad length range of the original notes than the baselines. In contrast, AUG-PE produces much shorter and more narrowly distributed sequences, indicating a loss of structural richness. FastDP also generates relatively short sequences, with a sharp peak around 500 tokens. These deviations suggest that Term2Note is better able to preserve the structural properties of real clinical notes, which is crucial for downstream utility and realism in synthetic data.

\paragraph{DP Quality Maximiser}

Beyond perplexity, we investigate a range of reference-free (RF) metrics to guide the selection of high-quality generations, including maximum and mean sentence length (in words and characters), self-BLEU, and distinct-$n$ variants. To evaluate these metrics, we manually annotate a small set of synthetic sections as ``good'' or ``bad'' based on readability, with  approximately 12\% labelled as ``bad''. Metrics are evaluated on their ability to identify these poor-quality sections via scalar thresholds. We then assess how well each metric identifies poor-quality sections using scalar thresholds. Our results indicate that metrics based on sentence length—particularly maximum sentence character count—align most closely with human annotations. A rejection threshold of 2181 characters yields the strongest correspondence.
Table~\ref{tab:text_distribution_stats_rf} reports the KL divergence and MAUVE values of synthetic notes after integrating this metric into the inference process. Specifically, if a generated section exceeds the threshold, it is discarded and regenerated until acceptance. Incorporating this simple criterion yields measurable improvements, suggesting that lightweight, reference-free filters can enhance the realism of DP synthetic text. Future work may extend this approach by combining multiple RF metrics for greater robustness.

\begin{table}
    \centering
    \begin{tabular}{lcc}
        \hline
        \textbf{Dataset} & \textbf{KL Divergence$\downarrow$} & \textbf{MAUVE$\uparrow$} \\
        \hline
        w/o RF metric & 1.99±0.37 & 0.24±0.04 \\
        w/ RF metric & 1.03±0.10 & 0.36±0.07 \\
        \hline
    \end{tabular}
    \caption{Fidelity evaluation of synthetic notes generated with and without integrating the RF metric (maximum sentence character count) into the inference process.
    }
    \label{tab:text_distribution_stats_rf}
\end{table}

\subsection{Utility}
Table \ref{tab:utility_result_sup} presents the detailed precision and recall scores for the downstream task evaluation.

\begin{table*}
    \centering
    \resizebox{\textwidth}{!}{
    \setlength{\tabcolsep}{1mm} 
    \begin{tabular}{ll cc l cc l cc l cc l cc}
        \toprule
        & \multirow{2}{*}{\textbf{Method}}
        & \multicolumn{2}{c}{\textbf{F1}}  &
        & \multicolumn{2}{c}{\textbf{Precision}}  &
        & \multicolumn{2}{c}{\textbf{Recall}}  &
        & \multicolumn{2}{c}{\textbf{AUC}}  &
        & \multicolumn{2}{c}{\textbf{Precision@$k$}} \\
        \cmidrule{3-4}\cmidrule{6-7}\cmidrule{9-10}
        \cmidrule{12-13}\cmidrule{15-16}
        & 
        & \textbf{Micro} & \textbf{Macro} &
        & \textbf{Micro} & \textbf{Macro} &
        & \textbf{Micro} & \textbf{Macro} &
        &  \textbf{Micro} & \textbf{Macro} &
        &  \textbf{$k=3$} & \textbf{$k=5$}  \\
        \midrule
        & {Original Data}  
        & 57.03±3.59 & 30.80±2.20  & & 60.14±4.51 & 34.66±4.56 & & 54.40±4.26 & 30.88±2.66 & & 82.01±1.36 & 58.88±2.73  & & 68.93±4.53 & 62.14±3.20 \\
        \midrule
        \multirow{32}{*}{\rotatebox{90}{Synthetic}} 
        & \multicolumn{15}{c}{$\epsilon=\infty$} \\ 
        \cmidrule(lr{0.001pt}){2-16}
        & AUG-PE $(\epsilon_\text{meta}=\infty, \epsilon_n=\infty)$
        & 45.82±2.33 & 14.84±2.97 & & 67.30±5.10  & 17.91±5.33 & & 34.94±3.19 & 16.27±2.67 & & 79.52±1.23  & 54.35±1.75 & & 68.48±6.11 & 61.77±3.03 \\
        & Fast-DP $(\epsilon_\text{meta}=\infty, \epsilon_n=\infty)$
        & 53.02±3.03 & 25.51±1.92 & & 56.19±5.62 & 26.83±2.31 & & 50.99±6.67 & 27.27±4.23 & & 79.35±1.55 & 51.05±2.48 & & 69.77±4.99 & 60.39±4.44 \\
        \addlinespace
        & Term2Note $(\epsilon_\text{meta}=\infty, \epsilon_n=\infty)$
        & 49.95±4.77  & 21.89±3.90 & & 65.37±6.31 & 28.69±1.26 & & 41.02±6.64  & 21.05±3.86 & & 81.40±1.78 & 55.43±2.40 & & 69.77±3.52 & 61.96±5.70 \\
        & \multicolumn{1}{l}{Term2Note $(\epsilon_t=\infty, \epsilon_n=\infty$)}
        & 49.24±2.63  & 21.9±2.01 & & 61.07±4.02 & 26.75±2.9 & & 41.61±4.73  & 21.64±3.29 & & 80.24±1.56 & 51.54±1.81 & & 67.81±3.66 & 61.08±3.71 \\
        & \multicolumn{1}{l}{Term2Note $(\epsilon_t=\infty, \epsilon_n=8)$}
        & 48.16±4.39 & 20.63±3.18 & & 59.07±4.04 & 26.64±5.14 & & 40.84±5.3 & 20.68±3.41 & &  78.72±0.85  & 50.01±4.19 & & 65.35±5.21 & 58.72±3.9   \\
        & \multicolumn{1}{l}{Term2Note $(\epsilon_t=\infty, \epsilon_n=5)$}
        & 51.0±1.41 & 22.73±2.22 & & 56.7±4.95 & 24.89±4.24 & & 46.79±3.93 & 24.62±3.03 & & 78.8±1.78 & 50.37±1.78 & & 67.15±5.88 & 59.91±3.12 \\
        & \multicolumn{1}{l}{Term2Note $(\epsilon_t=\infty, \epsilon_n=2)$}
        & 48.57±0.89 & 20.31±1.25 & & 59.92±5.19 & 24.63±2.69 & & 41.23±3.46 & 20.75±2.05 & & 79.56±1.08 & 51.75±0.85 & & 68.64±3.55 & 60.41±3.43 \\
        \addlinespace
        \cmidrule(lr{0.001pt}){2-16}
        & \multicolumn{15}{c}{$\epsilon=8$} \\
        \cmidrule(lr{0.001pt}){2-16}
        & AUG-PE $(\epsilon_\text{meta}=\infty, \epsilon_n=8)$
        & 40.73±9.22  & 13.28±4.71 & & 60.68±4.70 & 15.14±4.77 & & 31.43±9.80 & 16.01±5.81 & & 78.13±1.85 & 53.49±3.08 & & 63.24±4.65  & 59.03±4.59 \\
        & FastDP $(\epsilon_\text{meta}=\infty, \epsilon_n=8)$
        & 48.58±5.93 & 16.40±4.01 & & 64.79±1.33 & 16.70±3.67 & & 39.49±8.78 & 18.89±5.16 & & 80.74±1.41 & 51.57±3.46 & & 69.79±2.25 & 61.59±4.42 \\
        \addlinespace
        & Term2Note $(\epsilon_\text{meta}=\infty, \epsilon_n=8)$
        & 49.71±1.90  & 21.28±1.28  & & 61.37±4.28 & 25.17±1.55 & & 41.96±2.98 & 21.31±1.63 & & 80.03±1.44 & 52.80±1.15 & & 67.49±4.42 & 61.48±4.45 \\
        & \multicolumn{1}{l}{Term2Note $(\epsilon_t=8, \epsilon_n=8)$}
        & 52.31±4.45 & 26.5±3.95 & & 53.04±3.34 & 26.9±2.62 & & 51.87±6.45  & 28.66±4.65 & & 78.19±0.91 & 50.17±2.33 & & 67.81±5.88 & 57.36±3.3  \\
        & \multicolumn{1}{l}{Term2Note $(\epsilon_t=5, \epsilon_n=8)$}
        & 49.52±4.44 & 21.36±3.11 & & 58.74±6.32 & 24.59±5.36 & & 43.23±5.68 & 22.04±3.52 & & 79.08±1.34 & 50.11±4.08 & & 68.29±5.36 & 60.68±4.65 \\
        & \multicolumn{1}{l}{Term2Note $(\epsilon_t=2, \epsilon_n=8)$}
        & 53.25±1.1 & 24.66±1.82 & & 56.93±2.49 & 28.23±4.52 & & 50.16±2.52  & 26.94±2.17 & & 78.85±0.98 & 49.41±2.86 & & 68.31±5.53 & 59.91±3.34 \\
        \addlinespace
        \cmidrule(lr{0.001pt}){2-16}
        & \multicolumn{15}{c}{$\epsilon=5$} \\
        \cmidrule(lr{0.001pt}){2-16}
        & AUG-PE $(\epsilon_\text{meta}=\infty, \epsilon_n=5)$
        & 48.10±2.08 & 17.44±2.70 & & 60.59±8.29 & 18.28±3.69 & & 40.73±5.59 & 21.02±4.88 & & 77.78±3.11 & 53.33±2.60 & & 63.01±12.42 & 56.94±5.20 \\
        & Fast-DP $(\epsilon_\text{meta}=\infty, \epsilon_n=5)$
        & 49.30±4.04 & 16.23±3.04 & & 64.49±6.50 & 15.57±2.73 & & 40.84±7.33 & 19.70±4.87 & & 80.54±1.36 & 54.22±1.97 & & 67.31±7.22 & 61.98±3.34 \\
        \addlinespace
        & Term2Note $(\epsilon_\text{meta}=\infty, \epsilon_n=5)$
        & 47.94±4.47 & 20.31±3.40 & & 60.88±4.95 & 25.76±4.44 & & 40.15±6.22 & 20.20±3.71 & & 79.29±1.49 & 51.19±3.40 & & 66.04±5.73 & 61.69±4.86 \\
        & \multicolumn{1}{l}{Term2Note $(\epsilon_t=5, \epsilon_n=5)$}
        & 54.83±2.24 & 28.96±1.81 & & 51.64±4.44 & 29.07±4.58 & & 58.92±4.33 & 34.06±2.99 & & 78.2±0.8 & 50.32±3.94 & & 64.56±4.1 & 57.18±3.09  \\
        & \multicolumn{1}{l}{Term2Note $(\epsilon_t=2, \epsilon_n=5)$}
        & 51.26±1.97 & 21.45±1.88 & & 58.92±4.45 & 25.08±3.48 & & 45.63±3.55  & 23.28±1.99 & & 79.05±1.37 & 50.44±3.5 & & 66.36±5.89 & 60.49±4.89 \\
        \addlinespace
        \cmidrule(lr{0.001pt}){2-16}
        & \multicolumn{15}{c}{$\epsilon = 2$} \\
        \cmidrule(lr{0.001pt}){2-16}
        & AUG-PE $(\epsilon_\text{meta}=\infty, \epsilon_n=2)$
        & 40.9±7.43 & 13.57±3.88 & & 64.75±7.31 & 15.2±4.82 & & 30.7±9.22  & 14.83±5.29 & & 78.29±0.8 & 53.38±1.79 & & 63.74±5.0 & 60.1±4.37 \\
        & Fast-DP $(\epsilon_\text{meta}=\infty, \epsilon_n=2)$
        & 51.06±5.7 & 20.04±4.34 & & 58.78±5.13 & 19.77±4.27 & & 45.94±8.57  & 23.72±5.52 & & 79.98±1.95 & 51.31±3.32 & & 66.32±6.66 & 59.99±5.35 \\
        \addlinespace
        & Term2Note $(\epsilon_\text{meta}=\infty, \epsilon_n=2)$
        & 51.78±3.99 & 3.36±3.87 & & 57.45±5.97 & 25.59±4.84  & & 47.21±3.16 & 25.26±2.64 & & 79.00±1.67 & 50.60±2.75 & & 67.00±3.39 & 59.52±5.59 \\
        & \multicolumn{1}{l}{Term2Note $(\epsilon_t=2, \epsilon_n=2)$}
        & 51.87±2.73 & 23.06±2.51 & & 57.77±6.18 & 26.58±5.04 & & 47.37±2.88  & 24.77±2.01 & & 79.43±1.41 & 51.3±2.96 & & 69.45±4.73 & 60.31±5.39 \\
        \addlinespace
        \bottomrule
    \end{tabular}
    }
    \setlength{\tabcolsep}{6pt}  
    \caption{Supplementary results for utility evaluation: F1, Precision, Recall, AUC, and Precision@$k$, with \textbf{mean±standard deviation} values reported. 
    }
    \label{tab:utility_result_sup}
\end{table*}

\subsection{Gemma}
Experimental results for Gemma are reported in Table~\ref{tab:result_gemma}. Both fidelity and utility metrics are comparable to those of Llama in Table~\ref{tab:main_result}, although the MAUVE score for Gemma without DP (i.e., $\epsilon=\infty$) is higher than that of Llama. Overall, the same trend holds across both models: stricter privacy guarantees lead to reduced fidelity, while full privatisation still preserves strong fidelity and utility.

\begin{table*}
    \centering
    \resizebox{\textwidth}{!}{

    \setlength{\tabcolsep}{1mm} 
    \begin{tabular}{ll c l cc l c c| cc l cc l cc}
        \toprule
        & \multirow{3}{*}{\textbf{Method}} 
        & \multicolumn{6}{c}{\textbf{Fidelity}}
        & & \multicolumn{8}{c}{\textbf{Utility}} \\
        \cmidrule{3-8}\cmidrule{10-17}
        & 
        & \textbf{Length} 
        & 
        & \multicolumn{2}{c}{\textbf{Unary/Binary Term}} & 
        & \textbf{Semantic} &
        & \multicolumn{2}{c}{\textbf{F1}}  &
        & \multicolumn{2}{c}{\textbf{AUC}}  &
        & \multicolumn{2}{c}{\textbf{Precision@$k$}} \\
        \cmidrule{3-3}\cmidrule{5-6}\cmidrule{8-8}
        \cmidrule{10-11}\cmidrule{13-14}\cmidrule{16-17}
        & & \textbf{KL Div.}$\downarrow$ &
        & \textbf{Jaccard}$\uparrow$ & \textbf{KL Div.}$\downarrow$ & 
        & \textbf{MAUVE}$\uparrow$  &
        & \textbf{Micro} & \textbf{Macro} &
        &  \textbf{Micro} & \textbf{Macro} &
        &  \textbf{$k=3$} & \textbf{$k=5$}  \\
        \midrule
        & {Original Data} 
        & & & & & & & 
        & 57.03 & 30.80 & & 82.01 & 58.88 & & 68.93 & 62.14 \\
        \midrule
        \multirow{24}{*}{\rotatebox{90}{Synthetic}} 
        & \multicolumn{16}{c}{$\epsilon=\infty$} \\ 
        \cmidrule(lr{0.001pt}){2-17}
        & Term2Note $(\epsilon_\text{meta}=\infty, \epsilon_n=\infty)$
        & \textbf{0.18} & & \textbf{0.61/0.34} & \textbf{0.23}/1.55 & & \textbf{0.80} & 
        & 52.56 & 25.32 & & \textbf{80.81} & \textbf{55.09} & & 67.30 & 60.78 \\
        & \multicolumn{1}{l}{Term2Note $(\epsilon_t=\infty, \epsilon_n=\infty)$}
        & 0.36 & & 0.40/0.18 & 0.77/\textbf{1.52} & & 0.66 & 
        & 53.77 & 25.72 & & 79.94 & 51.86 & & 66.98 & 60.71 \\
        & \multicolumn{1}{l}{Term2Note $(\epsilon_t=\infty, \epsilon_n=8)$}
        & 0.40 & & 0.37/0.13 & 0.64/1.82 & & 0.38 & 
        & 47.63 & 20.19 & & 78.87 & 50.5 & & 66.83 & 59.53 \\
        & \multicolumn{1}{l}{Term2Note $(\epsilon_t=\infty, \epsilon_n=5)$}
        & 0.25 & & 0.37/0.13 & 0.65/1.76 & & 0.32 & 
        & 49.98 & 21.50 & & 79.77 & 52.57 & & 68.14 & 59.91 \\
        & \multicolumn{1}{l}{Term2Note $(\epsilon_t=\infty, \epsilon_n = 2)$}
        & 0.31 & & 0.36/0.13 & 0.69/1.59 & & 0.35 & 
        & 48.11 & 20.87 & & 79.49 & 52.83 & & 65.51 & 57.17 \\
        \addlinespace
        \cmidrule(lr{0.001pt}){2-17}
        & \multicolumn{16}{c}{$\epsilon=8$} \\
        \cmidrule(lr{0.001pt}){2-17}
        & Term2Note $(\epsilon_\text{meta}=\infty, \epsilon_n=8)$
        & 0.40 & & 0.39/0.13 & 0.53/1.66 & & 0.48 & 
        & 48.41 & 20.89 & & 78.55 & 48.89 & & 67.50 & 59.31 \\
        & \multicolumn{1}{l}{Term2Note $(\epsilon_t=8, \epsilon_n=8)$}
        & 0.40 & & 0.37/0.13 & 0.75/2.05 & & 0.32 & 
        & \textbf{55.46} & 26.73 & & 79.68 & 49.92 & & 69.30 & 60.60 \\
        & \multicolumn{1}{l}{Term2Note $(\epsilon_t=5, \epsilon_n=8)$}
        & 0.41 & & 0.38/0.13 & 0.70/1.71 & & 0.38 & 
        & 51.82 & 21.67 & & 80.44 & 51.28 & & \textbf{70.92} & 60.88 \\
        & \multicolumn{1}{l}{Term2Note $(\epsilon_t=2, \epsilon_n=8)$}
        & 0.51 & & 0.36/0.12 & 0.76/1.89 & & 0.30 & 
        & 52.34 & 22.50 & & 80.02 & 51.71 & & 69.62 & 60.01 \\
        \addlinespace
        \cmidrule(lr{0.001pt}){2-17}
        & \multicolumn{16}{c}{$\epsilon=5$} \\
        \cmidrule(lr{0.001pt}){2-17}
        & Term2Note $(\epsilon_\text{meta}=\infty, \epsilon_n=5)$
        & 0.36 & & 0.38/0.13 & 0.56/\textbf{1.52} & & 0.31 & 
        & 51.20 & 21.18 & & 80.03 & 53.10 & & 67.65 & \textbf{61.38} \\
        & \multicolumn{1}{l}{Term2Note $(\epsilon_t=5, \epsilon=5)$}
        & 0.39 & & 0.36/0.12 & 0.75/1.79 & & 0.32 & 
        & 52.79 & 23.74 & & 79.54 & 49.64 & & 67.99 & 58.92 \\
        & \multicolumn{1}{l}{Term2Note $(\epsilon_t=2, \epsilon=5)$}
        & 0.41 & & 0.36/0.13 & 0.77/1.83 & & 0.21 & 
        & 55.06 & \textbf{27.86} & & 79.42 & 49.19 & & 69.30 & 59.82 \\
        \addlinespace
        \cmidrule(lr{0.001pt}){2-17}
        & \multicolumn{16}{c}{$\epsilon = 2$} \\
        \cmidrule(lr{0.001pt}){2-17}
        & Term2Note $(\epsilon_\text{meta}=\infty, \epsilon_n=2)$
        & 0.49 & & 0.36/0.12 & 0.60/1.65 & & 0.27 & 
        & 49.26 & 21.17 & & 79.21 & 50.33 & & 67.16 & 59.24 \\
        & \multicolumn{1}{l}{Term2Note $(\epsilon_t=2, \epsilon_n=2)$}
        & 0.46 & & 0.35/0.12 & 0.80/1.78 & & 0.31 & 
        & 53.66 & 24.43 & & 79.91 & 50.26 & & 69.28 & 61.08 \\
        \addlinespace
        \bottomrule
    \end{tabular}
    }
    \setlength{\tabcolsep}{6pt}  
    \caption{Fidelity and utility evaluation of synthetic datasets generated by Term2Note with Gemma-3-1B as the base model for \textsc{NoteGen}. 
    }
    \label{tab:result_gemma}
\end{table*}

\section{Case Studies for Term Generation}
\label{appendix:term_generation_example}
Table~\ref{tab:example_terms} presents examples of synthetic clinical terms generated with and without the application of $\text{DPRP}^*$. The original list contains five salient terms extracted from a real clinical note. When no DP is applied, the generated list recovers only two of these terms (“air” and “discharge”), suggesting limited coverage despite the absence of privacy constraints. In contrast, the DP-enabled output does not directly replicate any of the original terms beyond “discharge”, but instead generates a substantially longer and more diverse list of medically plausible terms.

This illustrates a key trade-off: the DP mechanism introduces sufficient variability to obscure direct term recovery, thus enhancing privacy protection. At the same time, the generated list remains semantically coherent and clinically relevant, containing realistic phrases such as “hemodynamically stable,” “chronic low back pain,” and “pulmonary vein,” which contribute to the naturalness and utility of the resulting synthetic note. These findings align with our earlier quantitative analysis, confirming that $\text{DPRP}^*$ balances semantic fidelity with privacy-preserving diversity.

\begin{table*}
    \centering
    \begin{tabular}{p{0.15\textwidth}p{0.8\textwidth}}
        \toprule
        \textbf{Method} & \textbf{Terms} \\
        \midrule
        Original & [``Physical'', ``Discharge'', ``Laparoscopic'', ``incisions'', ``air''] \\
        \midrule
        No DP & [``air'', ``discharge''] \\
        \midrule
        DP & [``brief'', ``discharge'', ``negative'', ``medications'', ``placement'', ``drainage'', ``hemodynamically stable'', ``therapy'', ``chronic low back pain'', ``symptoms'', ``right chest'', ``referred to cardiac surgery'', ``chest discomfort'', ``pulmonary vein'', ``hyperlipidemia: he'', ``hypertension-'', ``difficulty'', ``surgical service'', ``anticoagulation'', ``discontinued'', ``increased'', ``afebrile'', ``asymptomatic'', ``admission'', ``intervention'', ``hospitalization'', ``cardiac enzymes x3''] \\
        \bottomrule
    \end{tabular}
    \caption{Example of synthetic terms.}
    \label{tab:example_terms}
\end{table*}

\end{document}